%% file: 2019-elsa-eusp.tex
\documentclass[compsoc,conference,a4paper,10pt,times,table]{IEEEtran}
\IEEEoverridecommandlockouts
\usepackage{cite}
\usepackage{amsmath,amssymb,amsfonts}
\usepackage[scaled=0.8]{beramono}
\usepackage{booktabs}
\usepackage{etoolbox}
\usepackage[T1]{fontenc}
\usepackage{graphicx}
\usepackage{listings}
\usepackage{microtype}
\usepackage[numbers,square,sort&compress]{natbib}
\usepackage{pifont}
\usepackage[binary-units=true]{siunitx}
\usepackage{soul}
\usepackage{subcaption}
\usepackage{textcomp}
\usepackage[nointegrals]{wasysym}
\usepackage{xcolor}
\usepackage{xspace}
\usepackage[colorlinks=true,citecolor=black,linkcolor=black,urlcolor=black]{hyperref}
\def\BibTeX{{\rm B\kern-.05em{\sc i\kern-.025em b}\kern-.08em
    T\kern-.1667em\lower.7ex\hbox{E}\kern-.125emX}}

\DeclareMathOperator*{\argmin}{arg\,min}
\newcommand{\perc}[1]{%
	\ifstrempty{#1}%
	{\,\si{\percent}}%
	{\SI{#1}{\percent}}%
}

\newcommand{\damd}{\textsc{Damd}\xspace}
\newcommand{\drebin}{Drebin+\xspace}
\newcommand{\mimicus}{Mimicus+\xspace}
\newcommand{\vuldee}{VulDeePecker\xspace}

\newcommand{\lemna}{LEMNA\xspace}
\newcommand{\lime}{LIME\xspace}
\newcommand{\lrp}{LRP\xspace}
\newcommand{\igLong}{Integrated Gradients\xspace}
\newcommand{\igShort}{IG\xspace}
\newcommand{\gradient}{Gradients\xspace}
\newcommand{\shap}{SHAP\xspace}

\newcommand{\Bbox}{Black-box\xspace}
\newcommand{\Wbox}{White-box\xspace}

\newcommand{\bbox}{black-box\xspace}
\newcommand{\wbox}{white-box\xspace}

\newcommand{\tab}[1]{Table~\hyperref[#1]{\ref{#1}}}
\newcommand{\fig}[1]{Figure~\hyperref[#1]{\ref{#1}}}
\newcommand{\figs}[1]{Figures~\hyperref[#1]{\ref{#1}}}
\newcommand{\sect}[1]{Section~\hyperref[#1]{\ref{#1}}}
\newcommand{\sects}[1]{Sections~\hyperref[#1]{\ref{#1}}}
\newcommand{\eqn}[1]{Equation~\hyperref[#1]{\ref{#1}}}
\newcommand{\appx}[1]{Section~\hyperref[#1]{\ref{#1}}}

\newcommand{\eg}{e.g.,\xspace} % ',' if American english else ''
\newcommand{\ie}{i.e.,\xspace} % ',' if American english else ''
\newcommand{\xv}{{x}}
\newcommand{\rv}{{r}}
\newcommand{\betav}{{\beta}}
\newcommand{\epsv}{{\epsilon}}
\newcommand{\piv}{{\pi}}
\newcommand{\code}[1]{\texttt{#1}}

\newcommand{\Cmark}{\ding{51}}%
\newcommand{\js}{JavaScript\xspace}

\newcommand{\tablesize}{\small}

\usepackage{amsthm}
\newtheoremstyle{ourstyle}
  {\topsep} % Space above
  {\topsep} % Space below
  {} % Body font
  {} % Indent amount
  {\bfseries} % Theorem head font
  {.} % Punctuation after theorem head
  {.5em} % Space after theorem head
  {} % Theorem head spec (can be left empty, meaning `normal')

\theoremstyle{ourstyle}
\newtheorem{defi}{Definition}
\makeatletter
\renewcommand\subsubsection{\@startsection {subsubsection}{3}{\z@}%
                                   {1ex plus 0.1ex minus 0.1ex}
                                   {0ex}%
                                   {\normalfont\bfseries}}

\makeatother
\DeclareMathOperator{\isop}{IS}
\newcommand{\is}{\ensuremath{\isop}\xspace}

\DeclareMathOperator{\daop}{DA}
\newcommand{\da}{\ensuremath{\daop}\xspace}

\DeclareMathOperator{\mazop}{MAZ}
\newcommand{\maz}{\ensuremath{\mazop}\xspace}

\definecolor{pro}{HTML}{60A7D2} % {2F7FBC} % {4F78A6} %{60BA6C}%{44FF77}
\definecolor{con}{HTML}{F18D2F} %{FF5544}
\definecolor{literal}{HTML}{4F78A6} %{1F77B4}
\definecolor{select}{HTML}{105BA4} %{309950}
\newcommand{\colorpro}{blue\xspace}
\newcommand{\colorcon}{orange\xspace}

\lstdefinestyle{mycppstyle}
{
	language=C++,
	keywordstyle=\bfseries,
	keywordstyle = [2]{\color{literal}},
	keywordstyle = [3]{\itshape},
	stringstyle=\color{literal},
	commentstyle=\color{gray},
	otherkeywords = {NULL, 50, 1, 00, wmemset, memmove},
	morekeywords = [2]{NULL, 50, 1, 00},
	morekeywords = [3]{wmemset, memmove},
}

\newcommand{\hlc}[2][yellow]{{%
		\colorlet{foo}{#1}%
		\sethlcolor{foo}\hl{#2}}%
}

\lstnewenvironment{codelst}[1][]{%
	\lstset{%
		escapeinside={(*@}{@*)},
		basicstyle=\footnotesize\ttfamily,
		breaklines=true,
		xleftmargin=5pt,
		framexleftmargin=5pt,
		frame=single,
		numbers=left,
		numberstyle=\tiny,
		belowskip=0em,
		#1
	}%
}{}

\newcommand{\webpage}{\texttt{http://explain-mlsec.org}}
\begin{document}

\title{Evaluating Explanation Methods\\ for Deep Learning in Security}
%\thanks{An earlier version of this paper has been
%	deposited on ArXiv.}  }

%\author{Anonymous submission}
\author{\IEEEauthorblockN{Alexander Warnecke\IEEEauthorrefmark{1}, Daniel Arp\IEEEauthorrefmark{1}, Christian Wressnegger\IEEEauthorrefmark{2} and Konrad Rieck\IEEEauthorrefmark{1}} \\
\IEEEauthorblockA{\IEEEauthorrefmark{1} Technische Universit\"at Braunschweig, Germany}
\IEEEauthorblockA{\IEEEauthorrefmark{2} Karlsruhe Institute of Technology, Germany}
%Brunswick, Germany}
%\and
}

\maketitle

\begin{abstract}
Deep learning is increasingly used as a building block of security
systems. Unfortunately, neural networks are hard to interpret and
typically opaque to the practitioner. The machine learning community
has started to address this problem by developing methods for
explaining the predictions of neural networks. While several of
these approaches have been successfully applied in the area of
computer vision, their application in security has received little
attention so far. It is an open question which explanation methods are
appropriate for computer security and what requirements they need to
satisfy.
In this paper, we introduce criteria for
comparing and evaluating explanation methods in the context of
computer security. These cover general properties, such as
the accuracy of explanations, as well as security-focused aspects, such
as the completeness, efficiency, and robustness. Based on our
criteria, we investigate six popular explanation methods and assess
their utility in security systems for malware detection and
vulnerability discovery.  We observe significant differences between
the methods and build on these to derive general recommendations for
selecting and applying explanation methods in computer security.
\end{abstract}

%\begin{IEEEkeywords}
%component, formatting, style, styling, insert
%\end{IEEEkeywords}

\section{Introduction}
Over the last years, deep learning has been increasingly recognized as
an effective tool for computer security.  Different types of neural
networks have been integrated into security systems, for example, for
malware detection~\citep{GroPapManBac+17, HuaSto16, McLRinKanYer+17},
\mbox{binary} analysis~\citep{ShiSonMoa15, ChuSheSaxLia+17,
XuLiuFenYin+17}, and vulnerability discovery \citep{LiZouXuOu+18}.
Deep learning, however, suffers from a severe drawback: Neural
networks are hard to interpret and their decisions are opaque to
practitioners. Even simple tasks, such as determining which features
of an input contribute to a prediction, are challenging to solve on
neural networks. This lack of transparency is a considerable problem
in security, as black-box learning systems are hard to audit and
protect from attacks \citep{CarWag17, PapMcDSinWel+18}.

The machine learning community has started to develop methods for
interpreting deep learning in computer vision
\mbox{\citep[\eg][]{ZeiFer14, BacBinMon+15, SimVedZis14}}. These
methods enable tracing back the predictions of neural networks to
individual regions in images and thereby help to understand the
decision process. These approaches have been further extended to also
explain predictions on text and sequences \citep[]{AraHorMon+17,
GuoMuXu+18}. Surprisingly, this work has received little attention in
security and there exists only a single technique that has been
investigated so
far~\citep{GuoMuXu+18}.

In contrast to other application domains of deep learning, computer
security poses particular challenges for the use of explanation
methods. First, security tasks, such as malware detection and binary
code analysis, require complex neural network architectures that are
challenging to investigate. Second, explanation methods in security do
not only need to be accurate but also satisfy security-specific
requirements, such as complete and robust explanations. As a result
of these challenges, it is an unanswered question which of the available
explanation methods can be applied in security and what properties they
need to possess for providing reliable results.

In this paper, we address this problem and develop evaluation criteria
for assessing and comparing explanation methods in security. Our work
provides a bridge between deep learning in security and explanation
methods developed for other application domains of machine
learning. Consequently, our criteria for judging explanations cover
general properties of deep learning as well as aspects that are
especially relevant to the domain of security.

\subsubsection*{General evaluation criteria}  As general criteria, we
consider the \emph{descriptive accuracy} and \emph{sparsity} of
explanations. These properties reflect how accurate and concise an
explanation method captures relevant features of a prediction. While
accuracy is an evident criterion for obtaining reliable results,
sparsity is another crucial constraint in security. In contrast to
computer vision, where an analyst can examine an entire image,
a security practitioner cannot investigate large sets
of features at once, and thus sparsity becomes an essential property
when non-graphic data is analyzed.

\subsubsection*{Security evaluation criteria} We define the
\emph{completeness}, \emph{stability}, \emph{robustness}, and
\emph{efficiency} of explanations as security criteria.  These
properties ensure that reliable explanations are available to a
practitioner in all cases and in reasonable time---requirements that
are less important in other areas of deep learning. For example, an
attacker may expose pathologic inputs to a security system that
mislead, corrupt, or slow down the computation of explanations. Note
that the robustness of explanation methods to adversarial examples is
not well understood yet, and thus we base our analysis on the recent
work by \citet{ZhaWanShe+19} and \citet{DomAlbAnd+19}.\\[-8pt]

With the help of these criteria, we analyze six recent explanation
methods and assess their performance in different security tasks. To
this end, we implement four security systems from the literature that
make use of deep learning and enable detecting Android
malware~\citep{GroPapManBac+17, McLRinKanYer+17}, malicious PDF
files~\citep{SrnLas14}, and security
vulnerabilities~\citep{LiZouXuOu+18}, respectively.
When explaining the decisions of these systems, we observe significant
differences between the methods in all criteria. Some methods are not
capable of providing sparse results, whereas others struggle with
structured security data or suffer from unstable outputs. While the
importance of the individual criteria depends on the particular task,
we find that the methods \igShort~\citep{SunTalYan17} and
\lrp~\citep{BacBinMon+15} comply best with all criteria and resemble
general-purpose techniques for security systems.

To demonstrate the utility of explainable learning, we also
qualitatively examine the generated explanations.  As an example for
this investigation, \fig{fig:vuldee-intro} shows three
explanations for the system \vuldee~\citep{LiZouXuOu+18} that
identifies vulnerabilities in source code. While the first explanation
method provides a nuanced representation of the relevant features, the
second method generates an unsharp explanation due to a lack of
sparsity. The third approach provides an explanation that even
contradicts the first one. Note that the variables \texttt{VAR2} and
\texttt{VAR3} receive a positive relevance (\colorpro) in the first
case and a negative relevance (\colorcon) in the third.

\begin{figure}[htbp] \centering
	\begin{subfigure}[t]{0.95\columnwidth} \input{tables/casestudy-vuldee-intro-code.tex}
		\label{fig:vuldee-code} \vspace{3mm}
	\end{subfigure}
	\begin{subfigure}[t]{0.95\columnwidth} \input{tables/intro-lrp.tex}
		\label{fig:vuldee-lemna} \vspace{3mm}
	\end{subfigure}
	\begin{subfigure}[t]{0.95\columnwidth} \input{tables/intro-lemna.tex}
		\label{fig:vuldee-lrp} \vspace{3mm}
	\end{subfigure}
	\begin{subfigure}[t]{0.95\columnwidth} \input{tables/intro-lime.tex}
		\label{fig:vuldee-lime}
	\end{subfigure}

	\caption{Explanations for the prediction of the security system
		\vuldee on a code snippet from the original dataset. From top to
		bottom: Original code, \lrp, \lemna, and \lime.}
	\label{fig:vuldee-intro}
\end{figure}

% Nicht: This analysis. Man denkt sonst an das Beispiel oben.
Our evaluation highlights the need for comparing explanation methods
and determining the best fit for a given security task. Furthermore,
it also unveils a notable number of artifacts in the underlying
datasets.
% Note that all examples in Figure~\ref{fig:vuldee-intro} highlight
% punctuation characters, such as a semicolons or commata, that are
% largely irrelevant for identifying vulnerabilities.
For all of the four security tasks, we identify features that are
unrelated to security but strongly contribute to the predictions. As a
consequence, we argue that explanation methods need to become an
integral part of learning-based security systems---first, for
understanding the decision process of deep learning and, second, for
eliminating artifacts in the training datasets.

The rest of this paper is organized as follows: We briefly review the
technical background of explainable learning in
Section~\ref{sec:explainable-learning}. The explanation methods and
security systems under test are described in Section
\ref{sec:methods}. We introduce our criteria for comparing explanation
methods in Section~\ref{sec:measures} and evaluate them in
Section~\ref{sec:quant}. Our qualitative analysis is presented in
Section~\ref{sec:insights-datasets} and Section~\ref{sec:conclusion}
concludes the paper.

\section{Explainable Deep Learning}
\label{sec:explainable-learning}

Neural networks have been used in artificial intelligence for over
\num{50}~years, yet concepts for explaining their decisions have just
recently started to be explored. This development has been driven by
the remarkable progress of deep learning in several areas, such as
image recognition~\citep{KriSutHin12} and machine
translation~\citep{SutVinLe14}. To embed our work in this context, we
briefly review two aspects of explainable learning that are crucial
for its application in security: the \emph{type of neural network
} and the \emph{explanation~strategy}.

\subsection{Neural Network Architectures}
\label{sec:architectures}

\begin{figure}[hbp]
	\centering \vspace{3mm}
	\begin{subfigure}[b]{0.3\columnwidth}
		\centering
		\includegraphics[height=22mm]{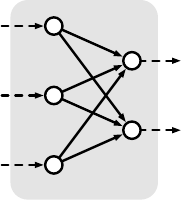}
		\subcaption{MLP layer}
	\end{subfigure}
	\begin{subfigure}[b]{0.3\columnwidth}
		\centering
		\includegraphics[height=22mm]{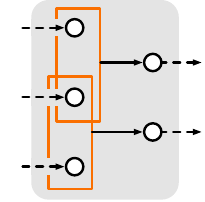}
		\subcaption{CNN layer}
	\end{subfigure}
	\begin{subfigure}[b]{0.3\columnwidth}
		\centering
		\includegraphics[height=22mm]{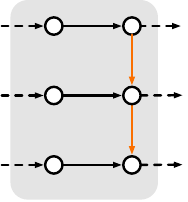}
		\subcaption{RNN layer}
	\end{subfigure}
	\caption{Overview of network architectures in security:
		Multilayer perceptrons (MLP), convolutional
		neural networks (CNN), and recurrent neural networks (RNN).}
	\label{fig:nn-overview}
\end{figure}

Different architectures can be used for constructing a neural network,
ranging from general-purpose networks to highly specific
architectures. In the area of security, three of these architectures
are prevalent: \emph{multilayer perceptrons}, \emph{convolutional
	neural networks}, and \emph{recurrent neural networks} (see
\fig{fig:nn-overview}).  Consequently, we focus our study on these
network types and refer the reader to the books by \citet{Roj96} and
\citet{GooBenCou16} for a detailed description of network
architectures in general.

\subsubsection{Multilayer Perceptrons (MLPs)} Multilayer perceptrons,
also referred to as \emph{feedforward networks}, are a classic and
general-purpose network architecture~\citep{RumGeoWil86}.  The network
is composed of multiple fully connected layers of neurons, where the
first and last layer correspond to the input and output of the
network, respectively. MLPs have been successfully applied to a
variety of security problems, such as intrusion and malware detection
\citep{GroPapManBac+17, HuaSto16}.  While MLP architectures are not
necessarily complex, explaining the contribution of individual
features is still difficult, as several neurons impact the decision
when passing through the network layers.

\subsubsection{Convolutional Neural Networks (CNNs)} These % neural
networks share a similar architecture with MLPs, yet they differ in
the concept of \emph{convolution} and \emph{pooling} \citep{LecBen95}.
The neurons in convolutional layers receive input only from a local
neighborhood of the previous layer. These neighborhoods overlap and
create receptive fields that provide a powerful primitive for
identifying spatial structure in data.  CNNs have thus been
successfully used for detecting malicious patterns in the bytecode of
Android applications \citep{McLRinKanYer+17}. Due to the convolution
and pooling layers, however, it is hard to explain the decisions of a
CNN, as its output needs to be ``unfolded'' and ``unpooled'' for
analysis.

\subsubsection{Recurrent Neural Networks (RNNs)} Recurrent networks,
such as LSTM and GRU networks~\citep{HocSch97, ChoMerBul+14}, are
characterized by a recurrent structure, that is, some neurons are
connected in a loop. This structure enables memorizing information and
allows RNNs to operate on sequences of data \citep{Elm90}. As a
result, RNNs have been successfully applied in security tasks
involving sequential data, such as the recognition of functions in
native code \citep{ShiSonMoa15, ChuSheSaxLia+17} or the discovery of
vulnerabilities in software~\citep{LiZouXuOu+18}. Interpreting the
prediction of an RNN is also difficult, as the relevance of an input
feature depends on the sequence of previously processed features.

\subsection{Explanation Strategies}
\label{sec:expl-strat}

Given the different architectures and the complexity of many neural
networks, decoding the entire decision process is a challenging task
that currently cannot be solved adequately.  However, there exist
several recent methods that enable explaining individual predictions
of a neural network instead of the complete decision
process~\citep[\eg][]{RibSinGue16, GuoMuXu+18, SunTalYan17, ZeiFer14,
	BacBinMon+15}. We focus on this form of explainable learning that
can be formally defined as follows:

\begin{defi} Given an input vector $\xv = (x_1,
\ldots, x_d)$, a neural network $N$, and a prediction $f_N(\xv) = y$,
an explanation method determines why the label~$y$ has been selected
by $N$.  This explanation is given by a vector $\rv = (r_1, \ldots,
r_d)$ that describes the relevance of the dimensions of $\xv$ for
$f_N(\xv)$.
\end{defi}

The computed relevance values $\rv$ are typically real numbers and can
be overlayed with the input in form of a heatmap, such that relevant
features are visually highlighted. An example of this visualization is
depicted in \fig{fig:vuldee-intro}. Positive relevance values are
shown in \colorpro and indicate importance \emph{towards} the
prediction $f_N(x)$, whereas negative values are given in \colorcon
and indicate importance \emph{against} the prediction. We will use
this color scheme throughout the paper\footnote{We use the
	\colorpro-\colorcon color scheme instead of the typical green-red
	scheme to make our paper better accessible to color-blind readers.}.

Despite the variety of approaches for computing a relevance vector for
a given neural network and an input, all approaches can be broadly
categorized into two explanation strategies: \emph{\bbox} and
\emph{\wbox} explanations.

\subsubsection{\Bbox Explanations} These methods operate under a \bbox
setting that assumes no knowledge about the neural network and its
parameters.  \Bbox methods are an effective tool if no access to the
neural network is available, for example, when a learning service is
audited remotely. Technically, \bbox methods rest on an approximation
of the function~$f_N$, which enables them to estimate how the
dimensions of~$\xv$ contribute to a prediction. Although \bbox methods
are a promising approach for explaining deep learning, they can be
impaired by the \bbox setting and omit valuable information provided
through the network architecture and parameters.

\subsubsection{\Wbox Explanations} These approaches operate under the
assumption that all parameters of a neural network are known and can
be used for determining an explanation.  As a result, these methods do
not rely on approximations and can directly compute explanations for
the function~$f_N$ on the structure of the network. In practice,
predictions and explanations are often computed from within the same
system, such that the neural network is readily available for
generating explanations. This is usually the case for stand-alone
systems for malware detection, binary analysis, and vulnerability
discovery. However, several \wbox methods are designed for specific
network layouts from computer vision and not applicable to all
considered architectures \citep[\eg][]{SimVedZis14, ZeiFer14,
	SprDosBro+15}.

% Until today, the majority of approaches from explainable learning
% has not been used for machine learning models in the security domain
% at all. In this paper, we examine multiple explanation algorithms
% from both classes and develop citeria to compare them. We will see
% that the methods create different explanations and introduce
% measures to determine which ones are best suited for the computer
% security domain.

% An intuitive way to visualize the results of explanation methods are
% heatmaps, where the relevance vector $\rv$ is visually overlayed with
% the input $\xv$. An example of this visualization is given in
% Figure~\fig{fig:vuldee-intro} for a sample from the \vuldee
% dataset. Positive relevances are shown in \colorpro and indicate
% importance \emph{towards} the decision of the classifier, whereas
% negative relevances are shown in \colorcon and indicate importance
% \emph{against} the classification.  We will use this color scheme
% throughout the paper.

% The code snippet contains no vulnerability and is classified
% correctly by the network. This means that the \code{strcpy} token
% speaks \emph{against} the classification meaning that it is
% indicative for a vulnerability while the \code{strlen} token speaks
% \emph{for} the classification, \ie indicating that no vulnerability
% is present here.

%\begin{figure}[tbp]
%  \centering \input{tables/intro-lrp.tex}
%  \caption{Explanation of a code snippet by \lrp.}
%  \label{fig:vuldee-lrp-2}
%\end{figure}

%
%\subsection{Relation to Other Concepts}

\Bbox and \wbox explanation methods often share similarities with
concepts of adversarial learning and feature selection, as these also
aim at identifying features related to the prediction of a
classifier. However, adversarial learning and feature selection pursue
fundamentally different goals and cannot be directly applied for
explaining neural networks. We discuss the differences to these
approaches for the interested reader in
Appendix~\ref{sec:relat-other-conc}.

\section{Methods and Systems under Test}
\label{sec:methods}

Before presenting our criteria for evaluating explanation methods, we
first introduce the methods and systems under test. In particular, we
cover six methods for explaining predictions in
Section~\ref{sec:explanation-methods} and present four security
systems based on deep learning in Section~\ref{sec:reference}.
For more information about explanation methods we do not evaluate in
the paper~\citep[\eg][]{DatSenSic16, FonVed17} we refer the reader to
the Appendix~\ref{sec:other-expl-meth}.

\subsection{Explanation Methods}
\label{sec:explanation-methods}

% There exist several \bbox and \wbox approaches for explaining the
% predictions of neural networks.
\tab{tab:overview-methods} provides an overview of popular
explanation methods along with their support for the different network
architectures. As we are interested in explaining predictions of
security systems, we select those methods for our study that are
applicable to all common architectures. In the following, we briefly
sketch the main idea of these approaches for computing relevance
vectors, illustrating the technical diversity of explanation methods.

\begin{table}[b]
	\centering\tablesize
	\caption{Popular explanation methods. The support
	for	different neural network architectures is
	indicated by \Cmark. Methods evaluated in this
	paper are indicated by \textcolor{select}{*}.}
	\begin{tabular}{l c@{\hspace{0.25cm}} c@{\hspace{0.25cm}} c}
		\toprule
		{\bfseries Explanation methods } &
		{\bfseries MLP } &
		{\bfseries CNN } &
		{\bfseries RNN }\\
		\cmidrule(lr){1-1}\cmidrule(lr){2-4}
		\textcolor{select}{Gradients*}~\citep{SimVedZis14},
		\textcolor{select}{IG*}~\citep{SunTalYan17}
		& \Cmark & \Cmark & \Cmark \\

		\textcolor{select}{{\lrp}*}~\citep{BacBinMon+15}, DeepLift~\citep{ShrGreKun17}
		& \Cmark & \Cmark & \Cmark \\

		PatternNet,
		PatternAttribution~\citep{KinSchAlb+18}
		& \Cmark & \Cmark &  --\\

		DeConvNet~\citep{ZeiFer14},
		GuidedBP~\citep{SprDosBro+15}
		& \Cmark & \Cmark &  --\\

		CAM~\citep{ZhoKhoLap+16},
		GradCAM~\citep{SelCogDas+17, ChaSarHow+18}
		& \Cmark & \Cmark &  --\\

		RTIS~\citep{DabGal17}, MASK~\citep{FonVed17}
		& \Cmark & \Cmark &  --\\

		\textcolor{select}{{\lime}*}~\citep{RibSinGue16},
		\textcolor{select}{{\shap}*}~\citep{LunLee17},
		QII~\citep{DatSenSic16}
		& \Cmark & \Cmark &  \Cmark\\

		\textcolor{select}{{\lemna}*}~\citep{GuoMuXu+18}
		& \Cmark & \Cmark & \Cmark \\

		\bottomrule
	\end{tabular}
	\label{tab:overview-methods}
\end{table}

\subsubsection{Gradients and \igShort}
One of the first \wbox methods to compute explanations for neural
networks has been introduced by \citet{SimVedZis14} and is based on
simple gradients.  The output of the method is given by
$r_i=\partial y/\partial x_i$, which the authors call a \emph{saliency
	map}. Here $r_i$ measures how much $y$ changes with respect to
$x_i$.  \citet{SunTalYan17} extend this approach and propose
\igLong(\igShort) that use a baseline~$x'$, for instance a vector of
zeros, and calculate the shortest path from $x'$ to $\xv$, given by
$\xv-x'$. To compute the relevance of $x_i$, the gradients with
respect to $x_i$ are cumulated along this path yielding
$$r_i = (x_i-x'_i)\int_0^{1}\frac{\partial f_N(x'+\alpha
	(x-x'))}{\partial x_i}\text{d}\alpha.$$
Both gradient-based methods can be applied to all relevant network
architectures and thus are considered in our comparative evaluation of
explanation methods.

\subsubsection{\lrp and DeepLift}
These popular \wbox methods determine the relevance of a prediction by
performing a backward pass through the neural network, starting at the
output layer and performing calculations until the input layer is
reached~\citep{BacBinMon+15}. The central idea of layer-wise relevance
propagation (\lrp) is the use of a conservation property that needs to
hold true during the backward pass. If $r_i^l$ is the relevance of the
unit~$i$ in layer~$l$ of the neural network then
$$\sum_i r_i^1 = \sum_i r_i^2 = \cdots = \sum_i r_i^L$$ needs to hold
true for all $L$~layers.
% further refine this property with the specific $\epsilon$-rule and
% $z$-rule.
%
Similarly, DeepLift performs a backward pass but takes a reference
activation $y'=f_N(x')$ of a reference input $\xv'$ into account. The
method enforces the conservation law,
$$\sum_i r_i = y-y'= \Delta y\,,$$ that is, the relevance
assigned to the features must sum up to the difference between the
outcome of $\xv$ and $\xv'$.
Both approaches support explaining the decisions of feed-forward,
convolutional and recurrent neural networks
\citep[see][]{AraHorMon+17}. However, as DeepLift and \igShort are
closely related~\citep{AncCeoOtr+18}, we focus our study on the method
$\epsilon$-\lrp.

% \begin{figure}[b]
%   \centering
%   \includegraphics[width=0.3\textwidth]{figs/lrp-overview}
%   \caption{Illustration of the explanation method \lrp.}
%   \vspace{-4mm}
%   \label{fig:lrp}
% \end{figure}

% \begin{figure}[b]
%   \centering
%   \includegraphics[width=0.3\textwidth]{figs/lemna-overview}
%   \caption{Illustration of the explanation method \lemna.}
%   \label{fig:lemna}
% \end{figure}

\subsubsection{\lime and \shap}
\citet{RibSinGue16} introduce one of the first \bbox methods for
explaining neural networks that is further extended by
\citet{LunLee17}. Both methods aim at approximating the decision
function $f_N$ by creating a series of $l$ perturbations of~$\xv$,
denoted as $\tilde{\xv}_1, \ldots, \tilde{\xv}_l$ by setting entries
in the vector~$\xv$ to \num{0} randomly. The methods then proceed by
predicting a label $f_N(\tilde{\xv}_i) = \tilde{y}_i$ for each
$\tilde{\xv}_i$ of the $l$ perturbations. This sampling strategy
enables the methods to approximate the local neighborhood of $f_N$ at
the point $f_N(\xv)$.
\lime \citep{RibSinGue16} approximates the decision boundary by a
weighted linear regression model,
$$
\argmin_{g\in\mathcal{G}} \sum_{i=1}^{l}
\pi_{\xv}(\tilde{\xv}_i)\big(f_N(\tilde{\xv}_i)-g(\tilde{\xv}_i)\big)^2,
$$
where $\mathcal{G}$ is the set of all linear functions and $\pi_x$ is
a function indicating the difference between the input $\xv$ and a
perturbation~$\tilde{\xv}$. \shap~\citep{LunLee17} follows the same
approach but uses the \shap kernel as weighting function $\pi_x$,
which is shown to create \emph{Shapley Values}~\citep{Sha53} when
solving the regression. Shapley Values are a concept from game theory
where the features act as players under the objective of finding a
fair contribution of the features to the payout---in this case the
prediction of the model. As both approaches can be applied to any
learning model, we study them in our empirical evaluation.

\subsubsection{\lemna}
As last explanation method, we consider \lemna, a \bbox method
specifically designed for security applications~\citep{GuoMuXu+18}.
It uses a mixture regression model for approximation, that is, a
weighted sum of $K$ linear models: % is used, which takes the form
$$ f(\xv) = \sum_{j=1}^K \piv_j(\betav_j \cdot \xv + \epsv_j).
$$
The parameter~$K$ specifies the number of models, the random variables
$\epsv=(\epsilon_1,\dots,\epsilon_K)$ originate from a normal
distribution $\epsilon_i \sim N(0,\sigma)$ and
$\piv=(\pi_1,\dots,\pi_K)$ holds the weights for each model.  The
variables $\betav_1,\dots,\betav_K$ are the regression coefficients
and can be interpreted as $K$~linear approximations of the decision
boundary near~$f_N(\xv)$.

\subsection{Security Systems}
\label{sec:reference}

As field of application for the six explanation methods, we consider
four recent security systems that employ deep
learning~(see~\tab{tab:overview}). The systems cover the three major
architectures/types introduced in \sect{sec:architectures} and comprise
between \num{4}~to~\num{6} layers of different types.

\begin{table}[htbp]\vspace{1mm}
	\caption{Overview of the considered security
	systems.}
	\centering\tablesize
	\begin{tabular}{
			llc
			S[table-format=1]
		}
		\toprule
		{\bfseries System} &
		{\bfseries Publication} &
		{\bfseries Type} &
		{\bfseries \# Layers} \\
		\cmidrule(lr){1-1}\cmidrule(lr){2-4}
		\drebin  & ESORICS'17~\citep{GroPapManBac+17} \hspace{-0mm} & MLP & 4 \\
		\mimicus & CCS'18~\citep{GuoMuXu+18}          & MLP & 4 \\
		\damd    & CODASPY'17~\citep{McLRinKanYer+17} \hspace{-0mm} & CNN & 6 \\
		\vuldee  & NDSS'18~\citep{LiZouXuOu+18}       & RNN & 5 \\
		\bottomrule
	\end{tabular}
	\label{tab:overview}\vspace{2mm}
\end{table}

\subsubsection{\drebin}
The first system %, called \drebin,
uses an MLP for identifying Android
malware. The system has been proposed by \citet{GroPapManBac+17} and
builds on features originally developed by \citet{ArpSprHueGasRie14}.
The network consists of two hidden layers, each comprising \num{200}
neurons.  The input features are statically extracted from Android
applications and cover data from the application's manifest, such as
hardware details and requested permissions, as well as information
based on the application's code, such as suspicious API calls and
network addresses.
To verify the correctness of our implementation, we train the system
on the original Drebin dataset~\citep{ArpSprHueGasRie14}, where we use
\perc{75} of the \num{129013}~Android application for training and
\perc{25} for testing. \tab{tab:performance} shows the results of this
experiment, which are in line with the performance published by
\mbox{\citet{GroPapManBac+17}}.

\subsubsection{\mimicus}
%.
The second system also uses an MLP but is designed to detect malicious
PDF documents. The system is re-implemented based on the work
of~\citet{GuoMuXu+18} and builds on features originally introduced by
\citet{SmuSta12}. Our implementation uses two hidden layers with
\num{200}~nodes each and is trained with \num{135}~features extracted
from PDF documents. These features cover properties about the document
structure, such as the number of sections and fonts in the document,
and are mapped to binary values as described by
\citet{GuoMuXu+18}. For a full list of features, we refer the reader
to the implementation by \citet{SrnLas14}.
%.
For verifying our implementation, we make use of the original dataset
that contains \num{5000}~benign and \num{5000}~malicious PDF files and
again split the dataset into \perc{75}~for training and \perc{25} for
testing. Our results are shown in \tab{tab:performance} and come close
to a perfect detection.

\subsubsection{\damd}
%.
The third security system studied in our evaluation
% is called \damd and
uses a CNN for identifying malicious Android
applications~\citep{McLRinKanYer+17}. The system processes the raw
Dalvik bytecode of Android applications and its neural network is
comprised of six layers for embedding, convolution, and max-pooling of
the extracted instructions. As the system processes entire
applications, the number of features depends on the size of the
applications. For a detailed description of this process, we refer the
reader to the publication by \mbox{\citet{McLRinKanYer+17}}. To
replicate the original results, we apply the system to data from the Malware
Genome Project~\citep{ZhoJia12}. This dataset consists of
\num{2123}~applications in total, with \num{863} benign and \num{1260}
malicious samples. We again split the dataset into \perc{75} of
training and \perc{25} of testing data and obtain results similar to
those presented in the original~publication.

\subsubsection{\vuldee}
%.
The fourth system uses an RNN for discovering vulnerabilities in
source code~\citep{LiZouXuOu+18}. The RNN consists of five layers,
uses \num{300}~LSTM cells~\citep{HocSch97}, and applies a word2vec
embedding~\citep{MikCheCor+13} with \num{200} dimensions for analyzing
C/C++ code. As a preprocessing step, the source code is sliced into
code gadgets that comprise short snippets of tokens. The gadgets are
truncated or padded to a length of \num{50}~tokens.
%
%  To avoid
% overfitting, identifiers and symbols are substituted with generic
% placeholders.
%
For verifying the correctness of our implementation, we use the
CWE-119 dataset, which consists of \num{39757} code gadgets, with
\num{10444} gadgets corresponding to vulnerabilities. In line with the
original study, we split the dataset into \perc{80}~training and
\perc{20}~testing data, and attain a comparable accuracy.

The four selected security systems provide a diverse view on the
current use of deep learning in security. \drebin and \mimicus are
examples of systems that make use of MLPs for detecting
malware. However, they differ in the dimensionality of the input:
While \mimicus works on a small set of engineered features, \drebin
analyzes inputs with thousands of dimensions. \damd is an example of a
system using a CNN in security and capable of learning from large
inputs, whereas \vuldee makes use of an RNN, similar to other
learning-based approaches analyzing program code
\citep[\eg][]{ShiSonMoa15, ChuSheSaxLia+17, XuLiuFenYin+17}.

\begin{table}
	\caption{Performance of the re-implemented
		security systems on the original datasets.}
	\centering\tablesize
	\begin{tabular}{
			l
			S[table-format=1.3]
			S[table-format=1.3]
			S[table-format=1.3]
			S[table-format=1.3]
		}
		\toprule
		{\bfseries System } &
		{\bfseries Accuracy} &
		{\bfseries Precision} &
		{\bfseries Recall} &
		{\bfseries F1-Score}\\
		%\midrule
		\cmidrule(lr){1-1}\cmidrule(lr){2-5}
		\drebin   & 0.980 & 0.926 & 0.924 & 0.925 \\
		\mimicus  & 0.994 & 0.991 & 0.998 & 0.994 \\
		\damd     & 0.949 & 0.967 & 0.924 & 0.953\\
		\vuldee\hspace{-3mm} & 0.908 & 0.837 & 0.802 &
		0.819\\
		\bottomrule
	\end{tabular}
	\label{tab:performance}

\end{table}

\section{Evaluation Criteria}
\label{sec:measures}

In light of the broad range of available explanation methods, the
practitioner is in need of criteria for selecting the best method for
a security task at hand. In this section, we develop these criteria
and demonstrate their utility in different examples. Before doing so,
however, we address another important question: Do the considered
explanation methods provide different results? If the methods
generated similar explanations, criteria for their comparison would be
less important and any suitable method could be chosen~in~practice.

To answer this question, we investigate the top-$k$ features of the six
explanation methods when explaining predictions of the security systems.
That is, we compare the set $T_i$ of the $k$~features with the highest
relevance from method $i$ with the set~$T_j$ of the $k$~features with
the highest relevance from method~$j$. In particular, we compute the
\emph{intersection size}
\begin{equation}
\isop(i,j)=\frac{\lvert T_i\cap T_j\rvert}{k},
\label{eq:intersection}
\end{equation}
as a measure of similarity between the two methods. The intersection
size lies between \num{0} and \num{1}, where \num{0} indicates no
overlap and \num{1} corresponds to identical top-$k$ features.

% We average this value over all samples of a dataset to obtain the
% average intersection size between the explanation methods.

\begin{figure}[t]
	\centering
	\includegraphics[width=\columnwidth,trim={0 0 0 5mm},clip]{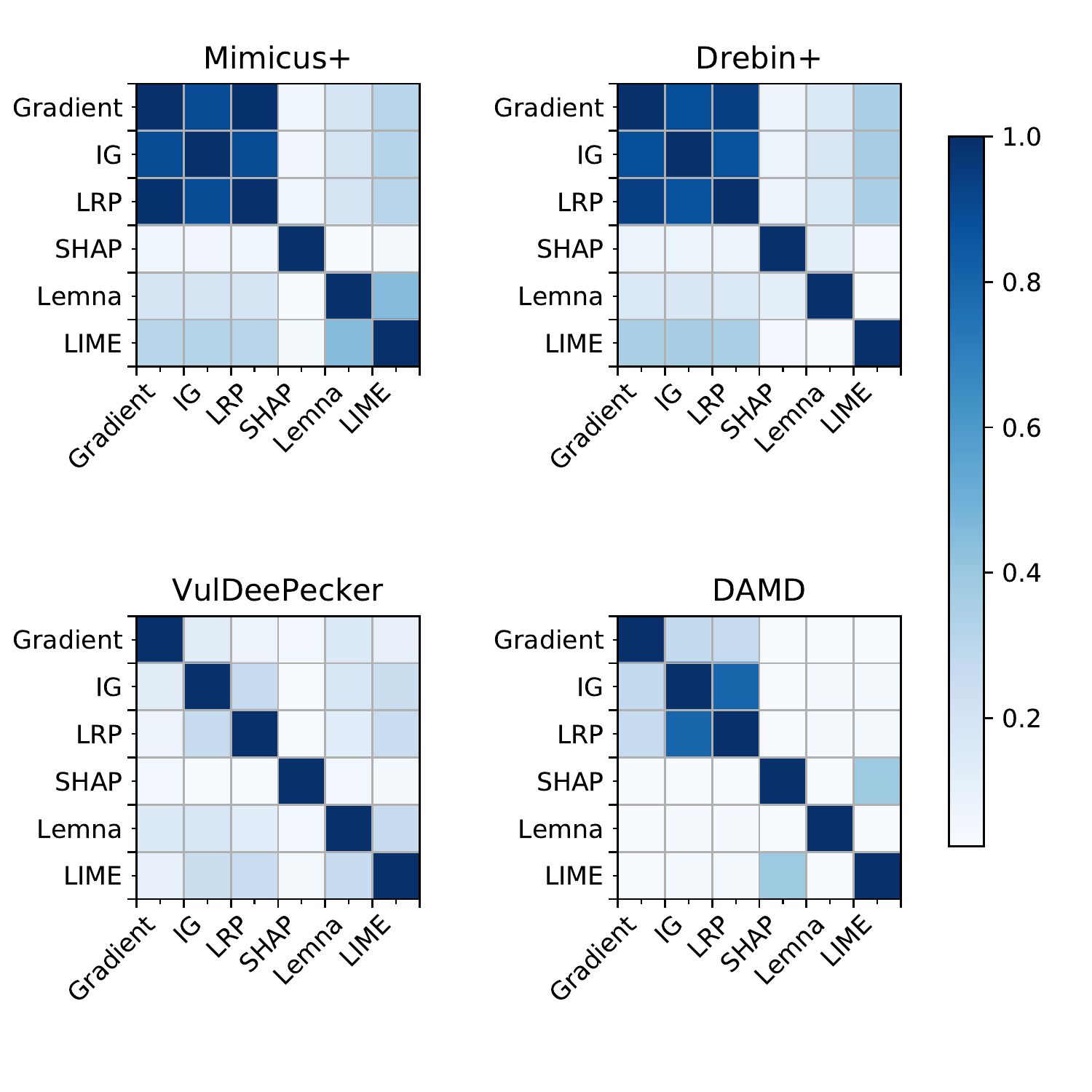}\vspace{-5mm}
	\caption{Comparison of the top-10 features for the different
		explanation methods. An average value of~\num{1} indicates
		identical top-10 features and a value of~\num{0} indicates
		no overlap.}\vspace{-2mm}
	\label{fig:methods_matrix}
\end{figure}

A visualization of the intersection size averaged over the samples of
the four datasets is shown in~\fig{fig:methods_matrix}.  We choose
$k=10$ according to a typical use case of explainable~learning: An
expert investigates the top-10 features to gain insights on a
prediction. For \damd, we use $k=50$, as the dataset is comprised of
long opcode sequences.  We observe that the top features of the
explanation methods differ considerably.  For example, in the case of
\vuldee, all methods determine different top-10 features. While we
notice some similarity between the methods, it becomes clear that the
methods cannot be simply interchanged, and there is a need for
measurable evaluation criteria.

% We find white and gray squares for all datasets under test
% indicating that only few top features are equal between the
% methods. We find that \gradient, \igShort and \lrp share great
% similarities for the \mimicus and \drebin dataset while being
% different to another for \vuldee. \lime is similar to \lemna for the
% \mimicus dataset and similar to \shap for the \damd dataset.

% \fig{fig:methods_matrix} raises the question which explanation method
% shall be used in practice. To answer this question we develop six
% quality criteria in the following subsections to allow a comparison
% between the approaches.

\subsection{General Criteria: Descriptive Accuracy}
\label{sec:intro-accuracy}
% The difference between the explanation algorithms
% in~\fig{fig:methods_matrix} raise the question for the \emph{best}
% important features that shall be investigated when analyzing a sample.

As the first evaluation criteria, we introduce the \emph{descriptive
	accuracy}. This criterion reflects how accurate an explanation
method captures relevant features of a prediction. As it is difficult
to assess the relation between features and a prediction directly, we
follow an indirect strategy and measure how removing the most relevant
features changes the prediction of the neural network.

\begin{defi} Given a sample $x$, the descriptive
accuracy (\da) is calculated by removing the $k$ most relevant features
$x_1,\dots,x_k$ from the sample, computing the new prediction using
$f_N$ and measuring the score of the original prediction class $c$
without the $k$ features,
$$\da_k\big(x, f_N\big) =
f_N\big(x\left.\right\rvert x_1=0,\dots,x_k=0\big)_c.$$
\end{defi}

%How is the descriptive accuracy interpreted?
% The best results are obtained when the softmax probability scores of
% $f_N$ are used, \ie $f_N(x)_c\in[0,1]$.
If we remove relevant features from a sample, the accuracy should
decrease, as the neural network has less information for making a
correct prediction. The better the explanation, the quicker the accuracy
will drop, as the removed features capture more context of the
predictions. Consequently, explanation methods with a steep decline of
the descriptive accuracy provide better explanations than methods with
a gradual decrease.

\begin{figure}[tbp]
	\begin{subfigure}[t]{\columnwidth}
		\input{tables/casestudy-vuldee-code.tex}
		\caption{Original code}
		\label{fig:case-study-vuldee-code}
	\end{subfigure}
	\begin{subfigure}[t]{\columnwidth}
		\input{tables/casestudy-vuldee-igShort.tex}
		\caption{\igLong}
		\label{fig:case-study-vuldee-lemna}
	\end{subfigure}
	\begin{subfigure}[t]{\columnwidth}
		\input{tables/casestudy-vuldee-lime.tex}
		\caption{\lime}
		\label{fig:case-study-vuldee-lrp}
	\end{subfigure}
	\caption{Explanations for a program slice from the
		\vuldee dataset using (b)~\igLong and (c)~\lime.}
	\label{fig:vuldee}
\end{figure}

\renewcommand{\line}[1]{%
	\mbox{line \num{#1}}\xspace
}
\newcommand{\lines}[3][ and ]{%
	\mbox{line \num{#2}#1\num{#3}\xspace}%
}
\newcommand{\linerange}[2]{%
	\lines[--]{#1}{#2}%
}

To demonstrate the utility of the descriptive accuracy, we consider a
sample from the \vuldee dataset, which is shown in
\fig{fig:vuldee}(a). The sample corresponds to a program slice and is
passed to the neural network as a sequence of
tokens. \figs{fig:vuldee}(b)~and~\ref{fig:vuldee}(c) depict these
tokens overlayed with the explanations of the methods \igLong
(\igShort) and \lime, respectively. Note that the \vuldee system
truncates all code snippets to a length of \num{50} tokens before
processing them through the neural network \citep{LiZouXuOu+18}.

The example shows a simple buffer overflow which originates from an
incorrect calculation of the buffer size in \line{7}. The two
explanation methods significantly differ when explaining the detection
of this vulnerability. While \igShort highlights the \code{wmemset}
call as important, \lime highlights the call to \code{memmove} and
even marks \code{wmemset} as speaking \emph{against} the detection.
Measuring the descriptive accuracy can help to determine which of
the two explanations reflects the prediction of the system better.

% originating from the definition of the fixed size buffers on
% \lines{2}{4}. Later on, in \line{7}, data from the one buffer,
% \code{source}, is copied to the other, \code{data}, using the
% \code{memmove} function. The third argument of the function
% determines the number of bytes that are copied, which is set to the
% size of \num{100}~wide~characters of type
% \code{wchar\_t}. Consequently, twice as many bytes are moved to the
% destination buffer than it may contain.

\subsection{General Criteria: Descriptive Sparsity}
\label{sec:intro-sparsity}

Assigning high relevance to features which impact a prediction is a
necessary prerequisite for good explanations. However, a human analyst
can only process a limited number of these features, and thus we
define the \emph{descriptive sparsity} as a further criterion for
comparing explanations methods as follows:

\begin{defi}
 The descriptive sparsity is measured
by scaling the relevance values to the range $[-1,1]$, computing a
normalized histogram $h$ of them and calculating the \emph{mass around
	zero} (\maz) defined by
$$\maz(r) = \int_{-r}^r h(x)\text{dx} ~\text{for}~ r\in[0,1].$$
\end{defi}

The \maz can be thought of as a window which starts at $0$ and grows
uniformly into the positive and negative direction of the $x$
axis. For each window, the fraction of relevance values that lies in the
window is evaluated. Sparse explanations have a steep rise in \maz
close to \num{0} and are flat around \num{1}, as most of the features
are not marked as relevant. By contrast, dense explanations have
a notable smaller slope close to \num{0}, indicating a larger set of
relevant features. Consequently, explanation methods with a \maz
distribution peaking at \num{0} should be preferred over methods with
less pronounced distributions.

\begin{table}[tbp]
	\caption{Explanations of \lrp and \lemna for a
		sample of the GoldDream family from the
		\damd~dataset.}
	\centering\tablesize
	\input{tables/casestudy-damd-shortened.tex}
	\label{tab:casestudy-damd}
\end{table}

As an example of a sparse and dense explanation, we consider two
explanations generated for a malicious Android application of the
\damd dataset. \tab{tab:casestudy-damd} shows a snapshot of these
explanations, covering opcodes of the \code{onReceive} method. \lrp
provides a crisp representation in this setting, whereas \lemna marks
the entire snapshot as relevant. If we normalize the relevance vectors
to $[-1,1]$ and focus on features above \num{0.2}, \lrp returns only
\num{14} relevant features for investigation, whereas \lemna returns
\num{2048} features, rendering a manual examination tedious.

% Moreover, when looking at the few highlighted tokens by \lrp we find
% the malicious behavior from the sample (see Appendix for an in-depth
% analysis of the sample).

It is important to note that the descriptive accuracy and the
descriptive sparsity are not correlated and must \emph{both} be
satisfied by an effective explanation method. A method marking all
features as relevant while highlighting a few ones can be accurate
but is clearly not sparse. Vice versa, a method assigning high
relevance to very few meaningless features is sparse but not accurate.

\begin{table*}[!htbp]
	\caption{Explanations for the Android malware FakeInstaller
		generated for \drebin using \gradient and \shap.}
	\centering\tablesize
	\input{tables/casestudy-drebin-incomplete.tex}
	\label{tab:drebin_incomplete}
\end{table*}

\subsection{Security Criteria: Completeness}
\label{sec:intro-completeness}

After introducing two generic evaluation criteria, we start focusing
on aspects that are especially important for the area of security. In a security
system, an explanation method must be capable of creating proper
results in all possible situations. If some inputs, such as
pathological data or corner cases, cannot be processed by an
explanation method, an adversary may trick the method into producing
\emph{degenerated} results. Consequently, we propose
\emph{completeness} as the first security-specific criterion.

\begin{defi}
An explanation method is complete, if
it can generate non-degenerated explanations for all possible input
vectors of the prediction function $f_N$.
\end{defi}

Several \wbox methods are complete by definition, as they calculate
relevance vectors directly from the weights of the neural network. For
\bbox methods, however, the situation is different: If a method
approximates the prediction function $f_N$ using random perturbations,
it may fail to derive a valid estimate of $f_N$ and return degenerated
explanations. We investigate this phenomenon in more detail in
Section~\ref{sec:availability}.

As an example of this problem, \tab{tab:drebin_incomplete} shows
explanations generated by the methods \gradient and \shap for a benign
Android application of the Drebin dataset.  The \gradient explanation
finds the \code{touchscreen} feature in combination with the
\code{launcher} category and the \code{internet} permission as an
explanation for the benign classification. \shap, however, creates an
explanation of zeros which provides no insights.
The reason for this degenerated explanation is rooted in the random
perturbations used by \shap. By flipping the value of features, these
perturbations aim at changing the class label of the input.  As there
exist far more benign features than malicious ones in the case of
\drebin, the perturbations can fail to switch the label and prevent
the linear regression to work resulting in a degenerated explanation.

% No matter which feature-combinations are set to zero for the
% perturbations, the sample will never be classified as malicious.
% The problem arises from the fact that only few features make a
% sample malicious, whereas there exists a large variety of benign
% features.  As a consequence, setting malicious features to~\num{0}
% for a perturbation usually leads to a benign classification. Setting
% benign features to zero, however, often does not impact the
% classification result. This leads to a set of perturbations with
% only one label thus the linear regression performed by \lime cannot
% create a meaningful result. The same holds true for \lemna and \shap
% since they are based on perturbations as well.

% To guarantee an explanation for every data sample we propose
% \emph{completeness} as a quality criteria for explanation methods.
% During our experiments with \num{500}~perturbations we find that the
% analyzed \bbox methods produce meaningful results if at least
% \num{20} of the perturbations (about \perc{4} on average) come from
% the opposite class. We thus analyze how often we find this amount of
% perturbations for the four considered security systems.

\subsection{Security Criteria: Stability}
\label{sec:intro-stability}
In addition to complete results, the explanations generated in a
security system need to be reliable. That is, relevant features must
not be affected by fluctuations and need to remain stable over time in
order to be useful for an expert. As a consequence, we define
\emph{stability} as another security-specific evaluation criterion.

\begin{defi}
 An explanation methods is stable, if
the generated explanations do not vary between multiple runs. That is,
for any run $i$ and $j$ of the method, the intersection size of the
top features $T_i$ and $T_j$ should be close to \num{1}, that is,
$\is(i, j) > 1 - \epsilon$ for some small threshold $\epsilon$.
\end{defi}

The stability of an explanation method can be empirically determined
by running the methods multiple times and computing the average
intersection size, as explained in the beginning of this
section. \Wbox methods are deterministic by construction since they
perform a fixed sequence of computations for generating an
explanation. Most \bbox methods, however, require random perturbations
to compute their output which can lead to different results for the
same~input.
%.
\tab{tab:mimicus_sample}, for instance, shows the output of \lemna for
a PDF document from the \mimicus dataset over two runs. Some of the
most relevant features from the first run receive very little
relevance in the second run and vice versa, rendering the explanations
unstable. We analyze these instabilities of the explanation methods in
Section~\ref{sec:stability}.

% Some \bbox methods suffer
% from this problem since the perturbations are often not
% representative for the dataset. For example, perturbations do not
% consider correlations between features which can lead to misleading
% results when predicting samples where only a few of those features
% are present.

\begin{table}
	\caption{Two explanations from \lemna for the
	same sample computed	in different runs.}
	\centering\tablesize
	\input{tables/casestudy-mimicus-no-stability-shortened.tex}
	\label{tab:mimicus_sample}
\end{table}

\subsection{Security Criteria: Efficiency}

When operating a security system in practice, explanations need to be
available in reasonable time. While low run-time is not a strict
requirement in general, time differences between minutes and
milliseconds are still significant. For example, when dealing with
large amounts of data, it might be desirable for the analyst to create
explanations for every sample of an entire class. We thus define
\emph{efficiency} as a further criterion for explanation methods in
security applications.

\begin{defi}
 We consider a method efficient if it
enables providing explanations without delaying the typical workflow
of an expert.
\end{defi}

As the workflow depends on the particular security task, we do not
define concrete run-time numbers, yet we provide a negative example as
an illustration.  The run-time of the method \lemna depends on the
size of the inputs. For the largest sample of the \damd dataset with
\num{530000} features, it requires about one hour for computing an
explanation, which obstructs the workflow of inspecting Android
malware severely.

\subsection{Security Criteria: Robustness}
\label{sec:intro-robustness}
As the last criterion, we consider the \emph{robustness} of
explanation methods to attacks. Recently, several
attacks~\citep[\eg][]{ZhaWanShe+19, DomAlbAnd+19, SlaHilJia19+} have
shown that explanation methods may suffer from adversarial
perturbations and can be tricked into returning incorrect relevance
vectors, similarly to adversarial examples \citep{CarWag17}. The
objective of these attacks is to disconnect the explanation from the
underlying prediction, such that arbitrary relevance values can be
generated that do not explain the behavior of the model.

\begin{defi}
 An explanation method is robust if the
computed relevance vector cannot be decoupled from the prediction by
an adversarial perturbation.
\end{defi}

Unfortunately, the robustness of explanation methods is still not well
understood and, similarly to adversarial examples, guarantees and
strong defenses have not been established yet. To this end, we
assess the robustness of the explanation methods based on the
existing literature. %While this strategy limits the
%evaluation of explanation methods in this study, this evaluation
%criterion can be easily adapted once a better understanding of attacks
%against explainable learning is established.

% Given a classification function $f$, an input $x$ the goal of an
% adversarial attack is to find an $\tilde{x}$ such that the difference
% between $x$ and $\tilde{x}$ is minimal but at the same time
% $f(x)\neq f(\tilde{x})$.  Given an explanation method $g$ that
% calculates relevances $g_f(x)$, \citet{ZhaWanShe+19} now also enforce
% that the explanations of $x$ and $\tilde{x}$ are similar by solving
% the adversarial attack

% \begin{equation}
% \begin{aligned}
% & \underset{\tilde{x}}{\text{min}}
% & & d_p\big(f(\tilde{x}), c_t\big) + \lambda d_e\big(g_f(\tilde{x}), g_f(x)\big) \\
% & \text{s.t. }
% & & \lvert\lvert x-\tilde{x}\rvert\rvert \leq \epsilon.\\
% \end{aligned}
% \label{eq:attack}
% \end{equation}

% \noindent where $d_p$ is the distance between the output classes of
% $f$, $c_t$ is the attacker's target class for $\tilde{x}$, $d_e$ is a
% distance function for explanations and $\lambda$ is a weighting
% parameter.

\section{Evaluation}
\label{sec:quant}

Equipped with evaluation criteria for comparing explanation methods,
we proceed to empirically investigate these in different security
tasks.  To this end, we implement a comparison framework that
integrates the six selected explanation methods and four security
systems.

\subsection{Experimental Setup}

\subsubsection{\Wbox Explanations}
For our comparison framework, we make use of the \code{iNNvestigate}
toolbox by \mbox{\citet{AlbLapSee+18}} that provides efficient
implementations for \lrp, \gradient, and \igShort. For the security
system \vuldee, we use our own \lrp
implementation~\citep{web:LRPForLSTMS} based on the publication
by~\citet{AraHorMon+17}. In all experiments, we set
\mbox{$\epsilon=10^{-3}$} for \lrp and use $N=64$ steps for
\igShort. Due to the high dimensional embedding space of \vuldee, we
choose a step count of $N=256$ in the corresponding experiments.

\subsubsection{\Bbox Explanations}
We re-implement \lemna in accordance to \citet{GuoMuXu+18} and use the
Python package \code{cvxpy}~\citep{DiaBoy16} to solve the linear
regression problem with Fused Lasso restriction~\citep{web:PaperRepo}.
We set the number of mixture models to \mbox{$K=3$} and the number of
perturbations to $l=500$. The parameter~$S$ is set to $10^4$ for
\drebin and \mimicus, as the underlying features are not sequential and
to $10^{-3}$ for the sequences of \damd and
\vuldee~\citep[see][]{GuoMuXu+18}. Furthermore, we implement \lime with
$l=500$ perturbations, use the cosine similarity as proximity measure,
and employ the regression solver from the \code{scipy} package using
$L_1$ regularization. For \shap we make use of the open-source
implementation by~\citet{LunLee17} including the~\code{KernelSHAP}
solver.

\newrobustcmd{\B}{\color{select}}
\newcommand{\spacing}{\hspace{0.3cm}}
\begin{table}[htbp]
	\caption{Descriptive accuracy (DA) and sparsity
		(MAZ) for the different explanation methods.}
	\begin{subfigure}[t]{\columnwidth}
		\centering\tablesize
		\newcommand{\mymidrule}{\cmidrule(lr){1-1}\cmidrule(lr){2-5}}
		\begin{tabular}{
				l@{\spacing}
				S[table-format=1.3]@{\spacing}
				S[table-format=1.3]@{\spacing}
				S[table-format=1.3]@{\spacing}
				S[table-format=1.3]@{\spacing}
			}
			\toprule
			{\bfseries Method } &
			{\bfseries \drebin} &
			{\bfseries \mimicus} &
			{\bfseries \damd} &
			{\bfseries \vuldee} \\
			%\midrule
			\mymidrule
			\lime      &  0.580 &  0.257 &  0.919 &  0.571 \\
			\lemna     &  0.656 &  0.405 &  0.983 &  0.764 \\
			\shap      &  0.891 &  0.565 &  0.966 &  0.869 \\
			\gradient  &  0.472 &  0.213 &  0.858 &  0.856 \\
			\igShort   &\B0.446 &\B0.206 &\B0.499 &\B0.574 \\
			\lrp       &  0.474 &  0.213 &  0.504 &  0.625 \\
			%			\mymidrule
			%			\baseline &  0.474 &  0.208 &  0.518 &  0.568 \\
			\bottomrule
		\end{tabular}
	    \vspace{1mm}
		\subcaption{Area under the DA curves from \fig{fig:aad}.}
		\label{tab:auc_remove}
	\end{subfigure}

	\begin{subfigure}[t]{\columnwidth}
		\vspace{2mm}
		\centering\tablesize
		\newcommand{\mymidrule}{\cmidrule(lr){1-1}\cmidrule(lr){2-5}}
		\begin{tabular}{
				l@{\spacing}
				S[table-format=1.3]@{\spacing}
				S[table-format=1.3]@{\spacing}
				S[table-format=1.3]@{\spacing}
				S[table-format=1.3]@{\spacing}
			}
			\toprule
			{\bfseries Method } &
			{\bfseries \drebin} &
			{\bfseries \mimicus} &
			{\bfseries \damd} &
			{\bfseries \vuldee} \\
			%\midrule
			\mymidrule
			\lime      &  0.757 &  0.752 &  0.833 &  0.745 \\
			\lemna     &  0.681 &  0.727 &  0.625 &  0.416 \\
			\shap      &  0.783 &  0.716 &  0.713 &  0.813 \\
			\gradient  &  0.846 &  0.856 &  0.949 &  0.816 \\
			\igShort   &\B0.847 &\B0.858 &\B0.999 &\B0.839 \\
			\lrp       &  0.846 &  0.856 &  0.964 &  0.827 \\
			%		\mymidrule
			%		\baseline &  0.849 &  0.857 &  0.998 &  0.846 \\
			\bottomrule
		\end{tabular}
	    \vspace{1mm}
		\subcaption{Area under \maz curves from \fig{fig:aad}.}
		\label{tab:auc_hist}
	\end{subfigure}
    \label{tab:conciseness}\vspace{-3mm}
\end{table}

\subsection{Descriptive Accuracy}
\label{sec:descriptive-accuracy}

We start our evaluation by measuring the descriptive accuracy~(DA) of
the explanation methods as defined in \sect{sec:intro-accuracy}. In
particular, we successively remove the most relevant features from the
samples of the datasets and measure the decrease in the classification
score. For \drebin and \mimicus, we remove features by setting the
corresponding dimensions to~$0$. For \damd, we replace the most relevant
instructions with the no-op opcode, and for \vuldee we substitute the
selected tokens with an embedding-vector of zeros.
% Moreover, we introduce a \baseline method as a baseline for this
% experiment. This method calculates the relevance $r_i$ by setting
% $x_i$ to zero and measuring the difference in the softmax
% probability outcome, i.e.  $r_i = f_N(x) - f_N(x|x_i=0)$.  We call
% this method \baseline because a sample with $d$ features has to be
% classified $d$ times again to calculate the relevances which can be
% time consuming for data with lots of features.

\begin{figure*}
	\centering
	\includegraphics[width=\textwidth]{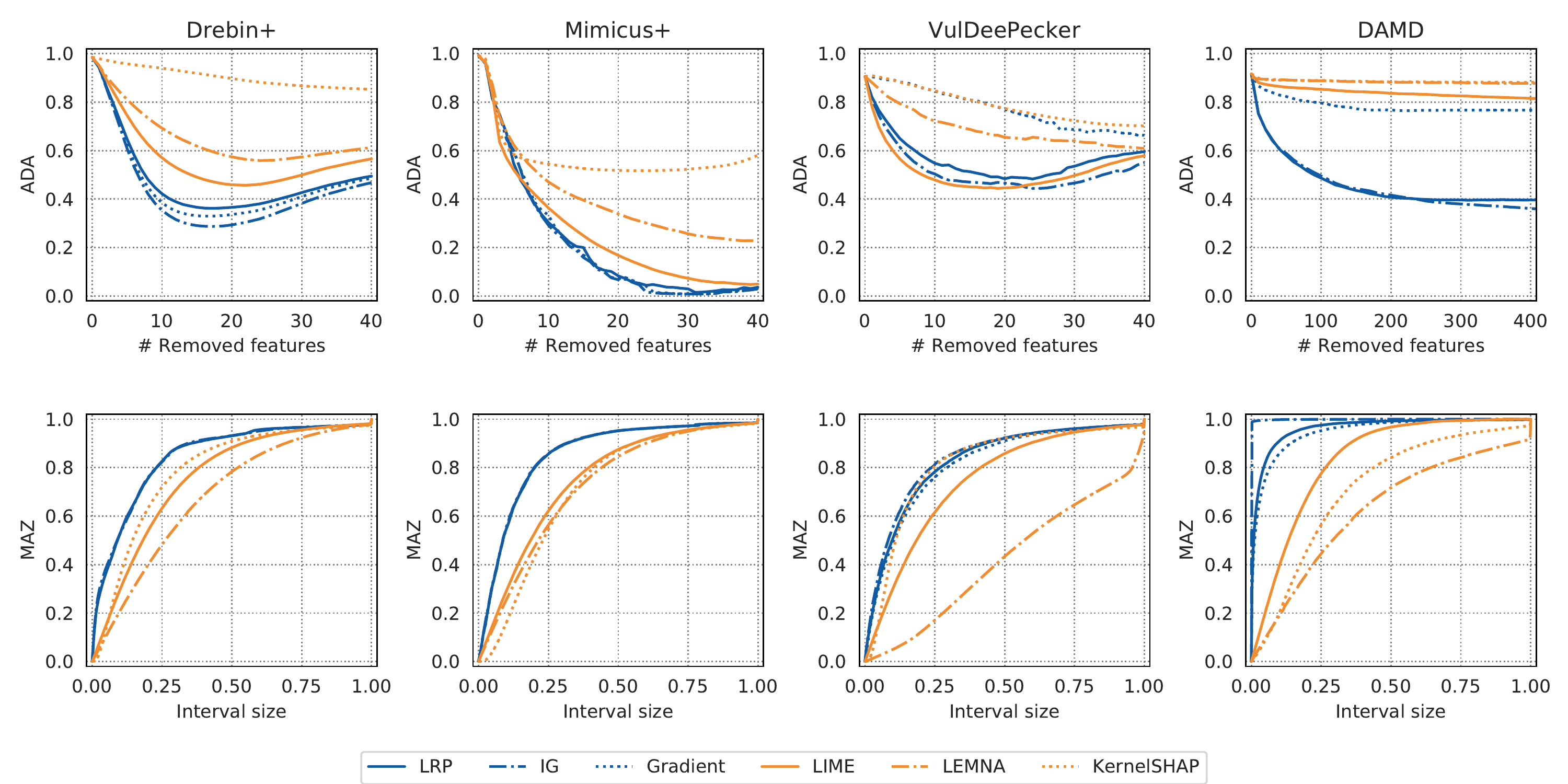}
	\caption
	{Descriptive accuracy and sparsity for the considered
	explanation methods.  Top row: Average descriptive accuracy
	(ADA); bottom row: sparsity measured as mass around zero (\maz).}
	\label{fig:aad}
\end{figure*}

The top row in \fig{fig:aad} shows the results of this experiment. As
the first observation, we find that the DA~curves vary significantly
between the explanation methods and security systems.  However, the
methods \igShort and \lrp consistently obtain strong results in all
settings and show steep declines of the descriptive accuracy. Only on
the \vuldee dataset, the \bbox method \lime can provide explanations
with comparable accuracy. Notably, for the \damd dataset, \igShort and
\lrp are the only methods to generate real impact on the outcome of the
classifier. For \mimicus, \igShort, \lrp and \gradient achieve a
perfect accuracy decline after only \num{25}~features and thus the
\wbox explanation methods outperform the \bbox methods in this
experiment.

Table~\ref{tab:conciseness}(a) shows the \emph{area under
curve}~(AUC) for the descriptive accuracy curves from~\fig{fig:aad}. We
observe that \igShort is the best method over all datasets---lower
values indicate better explanations---followed by \lrp. In comparison
to other methods it is up to \perc{48} better on average. Intuitively,
this considerable difference between the \wbox and \bbox methods makes
sense, as \wbox approaches can utilize internal information of the
neural networks that are not available to \bbox methods.

\subsection{Descriptive Sparsity}
\label{sec:sparsity}

We proceed by investigating the sparsity of the generated explanations
with the \maz score defined in \sect{sec:intro-sparsity}.
% .
The second row in \fig{fig:aad} shows the result of this experiment
for all datasets and methods. We observe that the methods \igShort,
\lrp, and \gradient show the steepest slopes and assign the majority
of features little relevance, which indicates a sparse
distribution. By~contrast, the other explanation methods provide flat
slopes of the \maz close to~$0$, as they generate relevance values
with a broader range and thus are less sparse.

For \drebin and \mimicus, we observe an almost identical level of sparsity
for \lrp, \igShort and \gradient supporting the findings from
\fig{fig:methods_matrix}.
Interestingly, for \vuldee, the \maz curve of \lemna shows a strong
increase close to~$1$, indicating that it assigns high relevance to a
lot of tokens. While this generally is undesirable, in case of \lemna,
this is founded in the basic design and the use of the Fused Lasso
constraint.
% .
In case of \damd, we see a massive peak at~$0$ for \igShort, showing
that it marks almost all features as irrelevant. According to the
previous experiment, however, it simultaneously provides a very good
accuracy on this data. The resulting sparse and accurate explanations
are particularly advantageous for a human analyst since the \damd
dataset contains samples with up to~\num{520000} features. The
explanations from \igShort provide a compressed yet accurate
representation of the sequences which can be inspected easily.

We summarize the performance on the \maz metric by calculating the
\emph{area under curve} and report it in \tab{tab:conciseness}(b).
A high AUC indicates that more features have been assigned a relevance
close to~$0$, that is, the explanation is more sparse. We find that the
best methods again are \wbox approaches, providing explanations that
are up to \perc{50} sparser compared to the other methods in this
experiment.

\subsection{Completeness of Explanations}
\label{sec:availability}

We further examine the completeness of the explanations. As shown in
\sect{sec:intro-completeness}, some explanation methods can not
calculate meaningful relevance values for all inputs. In particular,
perturbation-based methods suffer from this problem, since they
determine a regression with labels derived from random perturbations.
To investigate this problem, we monitor the creation of perturbations
and their labels for the different datasets.

\begin{figure}
	\centering
	\includegraphics[width=\columnwidth,trim={0 -4mm 0 2.5mm},clip]{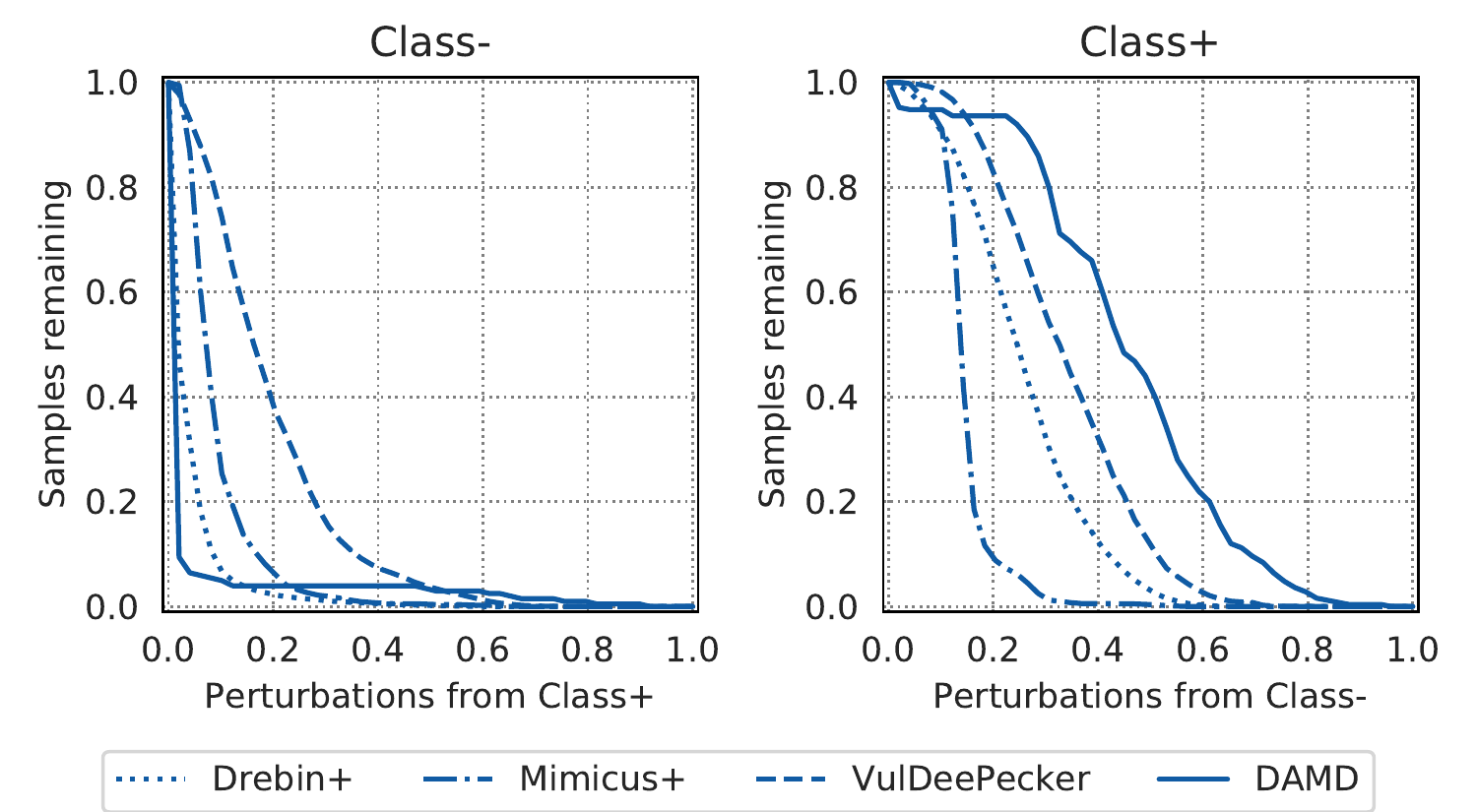}
	\caption
	{Perturbation label statistics of the datasets. For each percentage
	of perturbations from the other class the percentage of samples
    achieving this number is shown.}
	\label{fig:completeness}
\end{figure}

When creating perturbations for some sample $x$ it is essential for
\bbox methods that a fraction $p$ of them is classified as belonging to
the opposite class of $x$. In an optimal case one can achieve
$p\approx0.5$, however during our experiments we find that \perc{5} can
be sufficient to calculate a non-degenerated explanation in some cases.
\fig{fig:completeness} shows for each value of $p$ and all datasets the
fraction of samples remaining when enforcing a percentage $p$ of
perturbations from the opposite class.

In general, we observe that creating malicious perturbations from
benign samples is a hard problem, especially for \drebin and \damd. For
example, in the \drebin dataset only \perc{31} of the benign samples
can obtain a $p$ value of \perc{5} which means that more than \perc{65}
of the whole dataset suffer from degenerated explanations. A detailed
calculation for all datasets with a $p$ value of \perc{5} can be found
in \tab{tab:perturbations2} in the
Appendix~\ref{sec:completeness_example}.

The problem of incomplete explanations is rooted in the imbalance of
features characterizing malicious and benign data in the datasets.
While only few features make a sample malicious, there exists a large
variety of features turning a sample benign. As a consequence, randomly
setting malicious features to zero leads to a benign classification,
while setting benign features to zero usually does not impact the
prediction. As a consequence, it is often not possible to explain
predictions for benign applications and the analyst is stuck with an
empty explanation.

In summary, we argue that perturbation-based explanation methods
should only be used in security settings where incomplete explanations
can be compensated by other means. In all other cases, one should
refrain from using these \bbox methods in the context of security.

% For example, the smallest byte-sequence in the \damd dataset is
% eight bytes long and classified as benign. No matter which
% feature-combinations are set to zero for the perturbations, the
% sample will never be classified as malicious by the network.

\subsection{Stability of Explanations}
\label{sec:stability}

We proceed to evaluate the stability of the explanation methods when
processing inputs from the four security systems. To this end, we
apply the explanations to the same samples over multiple runs and
measure the average intersection size between the runs.

\tab{tab:stability} shows the average intersection size between the
top $k$ features for three runs of the methods as defined in
\eqn{eq:intersection}. We use $k=10$ for all datasets except for \damd
where we use $k=50$ due to the larger input space. Since the outputs
of \gradient, \igShort, and \lrp are deterministic, they reach the
perfect score of~$1.0$ in all settings and thus do not suffer from
limitations concerning stability.

For the perturbation-based methods, however, stability poses a severe
problem since none of those methods obtains a intersection size of
more than \num{0.5}. This indicates that on average half of the top
features do not overlap when computing explanations on the same input.
% The greatest deviation emerges for the \damd dataset where a
% potentially large number of perturbations is necessary to obtain a
% stable result for the high dimensional inputs.
Furthermore, we see that the assumption of \emph{locality} of the
perturbation-based methods does not apply for all models under test,
since the output is highly dependent on the perturbations used to
approximate the decision boundary. Therefore, the best methods for
the stability criterion beat the perturbation-based methods by a
factor of at least~\num{2.5} on all datasets.

\begin{table}[htbp]
	\caption{Average intersection size between top
	features for multiple runs. Values close to one
	indicate greater stability.}
	\centering\tablesize
	\newcommand{\mymidrule}{\cmidrule(lr){1-1}\cmidrule(lr){2-5}}
	\begin{tabular}{
			l@{\hspace{0.3cm}}
			S[table-format=1.3]@{\hspace{0.3cm}}
			S[table-format=1.3]@{\hspace{0.3cm}}
			S[table-format=1.3]@{\hspace{0.3cm}}
			S[table-format=1.3]@{\hspace{0.3cm}}
		}
		\toprule
		{\bfseries Method } &
		{\bfseries \drebin} &
		{\bfseries \mimicus} &
		{\bfseries \damd} &
		{\bfseries \vuldee} \\
		\mymidrule
		\lime      & 0.480    & 0.446   & 0.040    & 0.446  \\
		\lemna     & 0.4205    & 0.304   & 0.016    & 0.416  \\
		\shap      & 0.257    & 0.411   & 0.007    & 0.440  \\
		\gradient  & \B1.000     & \B1.000    & \B1.000     & \B1.000   \\
		\igShort   & \B1.000     & \B1.000    & \B1.000     & \B1.000   \\
		\lrp       & \B1.000     & \B1.000    & \B1.000     & \B1.000   \\
		%		\mymidrule
		%		\baseline  & \B1.000     & \B1.000    & \B1.000     & \B1.000   \\
		\bottomrule
	\end{tabular}
	\label{tab:stability}
	\vspace{-2mm}
\end{table}

\subsection{Efficiency of Explanations}
\label{sec:efficiency}

We finally examine the efficiency of the different explanation
methods. Our experiments are performed on a regular server system with
an~\code{Intel Xeon E5 v3} CPU at~\SI{2.6}{\giga\hertz}. It is
noteworthy that the methods \gradient, \igShort and \lrp can benefit
from computations on a graphical processing unit (GPU), therefore we
report both results but use only the CPU results to achieve a fair
comparison with the \bbox methods.

\tab{tab:runtime} shows the average run-time per input for all
explanations methods and security systems. We observe that \gradient
and \lrp achieve the highest throughput in general beating the other
methods by orders of magnitude. This advantage arises from the fact
that data can be processed \emph{batch-wise} for methods like
\gradient, \igShort, and \lrp, that is, explanations can be calculated
for a set of samples at the same time. The \mimicus dataset, for
example, can be processed in one batch resulting in a speed-up factor
of more than~\num{16000}$\times$ over the fastest \bbox
method. In general we note that the \wbox methods \gradient and \lrp
achieve the fastest run-time since they require a single backwards-pass
through the network. Moreover, computing these methods on a GPU results
in additional speedups of a factor up to three.
\renewcommand{\spacing}{\hspace{0.4cm}}
\begin{table}[!htbp]
	\caption{Run-time per sample in seconds. Note the range of the
	different times from microseconds to minutes.}
	\centering\tablesize\footnotesize
	\newcommand{\mymidrule}{\cmidrule(lr){1-1}\cmidrule(lr){2-5}}
	\resizebox{0.48\textwidth}{!}{
	\begin{tabular}{
		l@{\spacing}
		S[table-format=1.1e-1, table-align-exponent=true]@{\spacing}
		S[table-format=1.1e-1, table-align-exponent=true]@{\spacing}
		S[table-format=1.1e-1, table-align-exponent=true]@{\spacing}
		S[table-format=1.1e-1, table-align-exponent=true]
		}
		\toprule
		{\bfseries Method } &
		{\bfseries \drebin} &
		{\bfseries \mimicus} &
		{\bfseries \damd} &
		{\bfseries \vuldee} \\
		\mymidrule
		\lime  &  3.1e-2  &  2.8e-2  &  7.4e-1  &  3.0e-2 \\
		\lemna &  4.6e0   &  2.6e0   &  6.9e+2  &  6.1e0 \\
		\shap  &  9.1e0   &  4.3e-1  &  4.5     &  5.0e0 \\
		%\mymidrule
		\gradient &\B 8.1e-3 &\B  7.8e-6 &\B 1.1e-2  &\B 7.6e-4 \\
		\igShort  &   1.1e-1 &    5.4e-5 &   6.9e-1  &   4.0e-1 \\
		\lrp      &\B 8.4e-3 &\B  1.7e-6 &\B 1.3e-2  &   2.9e-2 \\
		\toprule
		{\bfseries GPU} &
		{\bfseries \drebin} &
		{\bfseries \mimicus} &
		{\bfseries \damd} &
		{\bfseries \vuldee} \\
		\mymidrule
		\gradient & 7.4e-3 &  3.9e-6 & 3.5e-3  & 3.0e-4 \\
		\igShort  & 1.5e-2 &  3.9e-5 & 3.0e-1  & 1.3e-1 \\
		\lrp      & 7.3e-3 &  1.6e-6 & 7.8e-3  & 1.1e-2 \\
		\bottomrule
	\end{tabular}}
	\label{tab:runtime}
\end{table}

The run-time of the \bbox methods increases for high dimensional
datasets, especially \damd, since the regression problems need to be
solved in higher dimensions. While the speed-up factors are already
enormous, we have not even included the creation of perturbations and
their classification, which consume additional run-time as well.

\subsection{Robustness of Explanations}
\label{sec:robustn-expl}

\newcommand{\good}{\textbf{+}\xspace}
\newcommand{\meh}{\textbf{\textasciitilde}\xspace}
\newcommand{\bad}{\textbf{--}\xspace}

\renewcommand{\good}{\CIRCLE\xspace}
\renewcommand{\meh}{\textcolor{lightgray}{\CIRCLE}\xspace}
\renewcommand{\bad}{\Circle\xspace}

\begin{table*}[!htbp]
	\caption{Results of the evaluated explanation
	methods. The last column summarizes these metrics
	in a rating comprising three levels: strong(\good),
	medium	(\meh), and weak (\bad).}
	\centering\tablesize
	\newcommand{\mymidrule}{\cmidrule(lr){1-1}\cmidrule(lr){2-7}\cmidrule(lr){8-8}}
	\begin{tabular}{
			l
			S[table-format=1.4]
			S[table-format=1.4]
			c
			S[table-format=1.4]
			S[table-format=1.1e-1, table-align-exponent=true,
			table-space-text-post=\,\si{\second}]
			c
			S
		}
		\toprule
		{\bfseries Explanation Method } &
		{\bfseries Accuracy} &
		{\bfseries Sparsity} &
		{\bfseries Completeness} &
		{\bfseries Stability} &
		{\bfseries Efficiency} &
		{\bfseries Robustness} &
		{\bfseries Overall Rating} \\
		\mymidrule
		\lime      & 0.582 & 0.772 &  -- & 0.353  &
		2.1e-1\,\si{\second} & \bad  &~\meh \meh \bad \bad \meh \bad
		\\
		\lemna     & 0.702 & 0.612 &  -- & 0.289  & 1.8e2
		\,\si{\second} & \bad &~\bad \bad \bad \bad \bad %\scalebox{1.5}[1.0]{--}\xspace\\
		\bad\\
		\shap      & 0.823 & 0.757 &  --   & 0.279 & 4.8e0
		\,\si{\second} & \bad & ~\bad \meh \bad \bad \bad \bad\\
		\gradient  & 0.600 & 0.867 & \Cmark & 1.000 &
		5.0e-3\,\si{\second} & \bad & ~\meh \good \good \good \good
		\bad\\
		\igShort   & 0.431 & 0.886 & \Cmark & 1.000 &
		3.0e-1\,\si{\second} & \bad & ~\good \good \good \good \meh
		\bad\\
		\lrp       & 0.454 & 0.873 & \Cmark & 1.000 &
		5.0e-2\,\si{\second} & \bad &~\good \good \good \good \good
		\bad\\
		%\mymidrule
		%\baseline & 0.4420 & 0.8875 & \Cmark & 1.0 & 1.30e2
		%\,\si{\second} & ? &~\good \good \good \bad \good \meh\\
		\bottomrule
	\end{tabular}
	\label{tab:summary}
\end{table*}

Recently, multiple authors have shown that adversarial perturbations are
also applicable against explanation methods and can manipulate the
generated relevance values.
% .
Given a classification function $f$, an input $x$ and a target class
$c_t$ the goal of an adversarial perturbation is to find
$\tilde{x}=x+\delta$ such that $\delta$ is minimal but at the same time
$f(\tilde{x})=c_t\neq f(x)$.

For an explanation method $g_f(x)$ \citet{ZhaWanShe+19} propose to
solve
\begin{equation}
\underset{\delta}{\text{min }}
d_p\big(f(\tilde{x}), c_t\big) + \lambda d_e\big(g_f(\tilde{x}), g_f(x)\big), \\
\label{eq:attack_zhang}
\end{equation}
\noindent where $d_p$ and $d_e$ are distance measures for classes and
explanations of $f$. The crafted input $\tilde{x}$ is misclassified by
the network but keeps an explanation very close to the one of $x$.
%However, they also show that the methods \gradient, \igShort, and \lrp
%are robust against their attack. \citet{GhoAbiZou19} show that
%indistinguishable adversarial samples $\tilde{x}$ can be crafted in
%such a way that they receive an explanation with a large deviation from
%the original one and methods like \igShort and \gradient are vulnerable
%to this attack.
\citet{DomAlbAnd+19} show that many \wbox methods can be
tricked to produce an arbitrary explanation $e_t$ without changing the
classification by solving

\begin{equation}
\underset{\delta}{\text{min }}
d_e\big(g_f(\tilde{x}), e_t\big) + \gamma d_p\big(f(\tilde{x}), f(x)\big).\\
\label{eq:attack_dombrowski}
\end{equation}

While the aforementioned attacks are constructed for \wbox methods,
\citet{SlaHilJia19+} have recently proposed an attack against
\lime and \shap. They show that the perturbations, which have to be
classified to create explanations, deviate strongly from the
original data distribution and hence are easily distinguishable from
original data samples. With this knowledge an adversary can use a
different model $\tilde{f}$ to classify the perturbations and create
arbitrary explanations to hide potential biases of the original model.
%They show that an attacker who controls the
%perturbations can make the explanations deviate largely from the true
%behavior of the classifier.
Although \lemna is not considered by~\citet{SlaHilJia19+}, it can be
attacked likewise since it relies on perturbation labels as well.

The \wbox attacks by~\citet{ZhaWanShe+19} and \citet{DomAlbAnd+19}
require access to the model parameters which is a technical hurdle in
practice. Similarly, however, the \bbox attack by \citet{SlaHilJia19+}
needs to bypass the classification process of the perturbations to
create arbitrary explanations which is equally difficult. A further
problem of all attacks in the security domain are the discrete input
features: For images, an adversarial perturbation $\delta$ is
typically small and imperceptible, while binary features, as in the
\drebin dataset, require larger changes with
$\lvert\delta\rvert\geq 1$. Similarly, for \vuldee and \damd, a direct
application of existing attacks will likely result in broken code or
invalid behavior. Adapting these attacks seems possible but requires
further research on adversarial learning in structured domains.

Based on this analysis, we conclude that explanation methods are not
robust and vulnerable to different attacks. Still, these attacks
require access to specific parts of the victim system as well as
further extenions to work in discrete domains. As a consequence, the
robustness of the methods is difficult to assess and further work is
needed to establish a better understanding of this threat.

\subsection{Summary}

A strong explanation method is expected to achieve good results for
each criterion and on each dataset. For example, we have seen that the
\gradient method computes sparse results in a decent amount of time.
The features, however, are not accurate on the \damd and \vuldee
dataset.
% .
Equally, the relevance values of \shap for the \drebin dataset are
sparser than those from \lemna but suffer from instability. To provide
an overview, we average the performance of all methods over the four
datasets and summarize the results in \tab{tab:summary}.

For each of the six evaluation criteria, we assign each method one of
the following three categories: \good, \meh, and~\bad. The
\good~category is given to the best explanation method and other
methods with a similar performance. The \bad~category is
assigned to the worst method and methods performing equally bad.
Finally, the \meh~category is given to methods that lie between the
best and worst methods.% and thus provide medium results.

Based on \tab{tab:summary}, we can see that \wbox explanation methods
achieve a better ranking than \bbox methods in all evaluation
criteria. Due to the direct access to the parameters of the neural
network, these methods can better analyze the prediction function and
are able to identify relevant features. In particular, \igShort and
\lrp are the best methods overall regarding our evaluation
criteria. They compute results in less than \SI{50}{\ms} in our
benchmark, mark only few features as relevant, and the selected
features have great impact on the decision of the classifier. These
methods also provide deterministic results and do not suffer from
incompleteness.  As a result, we recommend to use these methods for
explaining deep learning in security.
However, if \wbox access is not available, we recommend the \bbox
method \lime as it shows the best performance in our experiments or to
apply \emph{model stealing} as shown in the following
\sect{sec:model-stealing} to enable the use of \wbox methods.

In general, whether \wbox or \bbox methods are applicable also depends
on \emph{who} is generating the explanations: If the developer of a
security system wants to investigate its prediction, direct access to
all model parameters is typically available and \wbox methods can be
applied easily.  Similarly, if the learning models are shared between
practitioners, \wbox approaches are also the method of choice. If the
learning model, however, is trained by a remote party, such as a
machine-learning-as-a-service providers, only \bbox methods are
applicable. Likewise, if an auditor or security tester inspects a
proprietary system, \bbox methods also become handy, as they do not
require reverse-engineering and extracting model parameters.

%and up to our knowledge there exist no
%\bbox applications that allow an infinite amount of requests from a
%user.To this end, we regard the use-cases of \bbox methods as limited.
%Similarly, we argue that commercial products should provide \wbox
%explanations created by the vendor so that customers are not exposed to
%the limitations of \bbox methods when trying to understand the
%delivered predictions.

\subsection{Model Stealing for White-Box Explanations}
\label{sec:model-stealing}
Our experiments show that practitioners from the security
domain should favor \wbox methods over \bbox methods when aiming to
explain neural networks. However, there are cases when access to the
parameters of the system is not available and \wbox methods can not be
used. Instead of using \bbox methods one could also use \textit{model
stealing} to obtain an approximation of the original
network\citep{TraZhaJue16+}. This approach assumes that the user can
predict an unlimited number of samples with the model to be explained.
The obtained predictions can  then be used to train a surrogate model
which might have a different architecture but a similar behavior.

To evaluate the differences between the explanations of surrogate models
to the original ones we conduct an experiment on the \drebin and
\mimicus datasets as follows: We use the predictions of the original
model from~\citet{GroPapManBac+17} which has two dense layers with
$200$ units each and use these predictions to train three surrogate
models. The number of layers is varied to be $[1,2,3]$ and the number
of units in each layer is always $200$ resulting in models with higher,
lower and the original complexity. For each model we calculate
explanations via \lrp and compute the intersection size given by
\eqn{eq:intersection} for $k=10$.

\begin{figure}
	\centering
	\includegraphics[width=\columnwidth,clip]{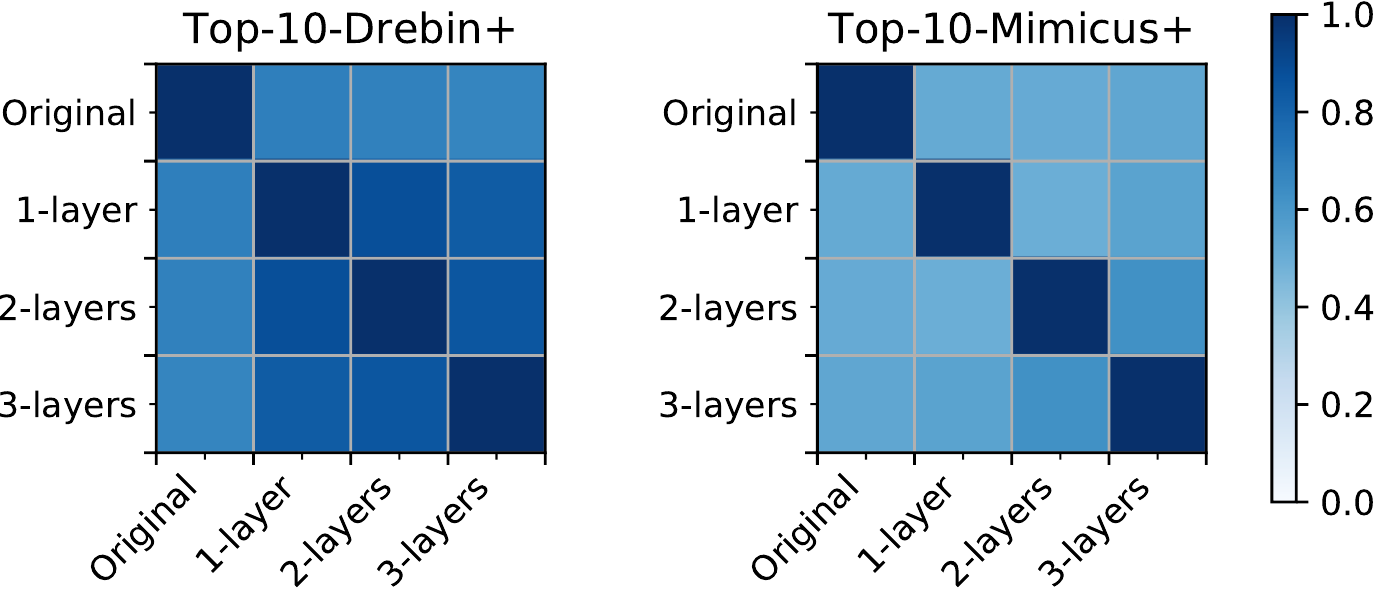}
	\caption{Intersection size of the Top-10 features of explanations
	obtained from models that were stolen from the original model of
	the \drebin and \mimicus dataset.}
	\label{fig:model_stealing_matrix}
\end{figure}

The results in \fig{fig:model_stealing_matrix} show that the models
deliver similar explanations to the original model
(IS$\approx$\num{0.7}) although having different architectures for
the \drebin dataset. However, the similarity between the stolen models
is clearly higher (IS$\approx$\num{0.85}). For the \mimicus dataset, we
observe a general stability of the learned features at a lower level
(IS$\approx$\num{0.55}). These results indicate that the explanations of
the stolen models are better than those obtained from \bbox
methods (see \fig{fig:methods_matrix}) but still deviate from the
original model, \ie there is no global transferability between the
explanations. At all, model stealing can be considered a good
alternative to the usage of \bbox explanation methods.

\section{Insights on the Datasets}
\label{sec:insights-datasets}

During the experiments for this paper, we have analyzed various
explanations of security systems---not only quantitatively as
discussed in~\sect{sec:quant} but also qualitatively from the
perspective of a security analyst. In this section, we summarize our
observations and discuss insights related to the role of deep learning
in security.
Moreover, we publish the generated explanations from all
datasets and methods on the project's website\footnote{\webpage} in
order to foster future research.

\subsection{Insights on \mimicus}

When inspecting explanations for the \mimicus system, we observe that
the features for detecting malware are dominated by
\code{count\_javascript} and \code{count\_js}, which both stand for
the number of \js elements in the document. The strong impact of these
elements is meaningful, as \js is frequently used in malicious PDF
documents~\cite{KapShoCovKruVig13}. However, we also identify features
in the explanations that are non-intuitive.  For example, features
like \code{count\_trailer} that measures the number of trailer
sections in the document or \code{count\_box\_letter} that counts the
number of US letter sized boxes can hardly be related to security and
rather constitute artifacts in the dataset captured by the learning
process.

To further investigate the impact of \js features on the neural
network, we determine the distribution of the top~\num{5} features from
the method \igShort for each class in the entire dataset. It turns out
that \js appears in \perc{88} of the malicious documents, whereas only
about~\perc{6} of the benign samples make use of it
(see~\tab{tab:mimicus_stats}). This makes \js an extremely
discriminating feature for the dataset. From a security perspective,
this is an unsatisfying result, as the neural network of \mimicus
relies on a few indicators for detecting the malicious code in the
documents.  An attacker could potentially evade \mimicus by not using
\js or obfuscating the \js elements in the document.

%\vspace{2mm}
\begin{table}
	\caption{Top-5 features for the \mimicus dataset
	determined using \igShort. The right columns
	show
	the frequency in benign and malicious PDF
	documents, respectively.}
	\centering\tablesize
	\begin{tabular}{
		c
		l
		S[table-format=3.1,table-space-text-post={\perc{}}]
		S[table-format=2.1,table-space-text-post={\perc{}}]
		}
		\toprule
		{\bfseries Class \hspace{-2.5mm}} &
		{\bfseries Top 5 Feature} &
		{\bfseries Benign} &
		{\bfseries Malicious }\\
		\midrule
		-- & \code{count\_font}  & 98.4\perc{} &
		20.8\perc{} \\
		-- & \code{producer\_mismatch}\hspace{-3mm} & 97.5\perc{} &
		16.6\perc{} \\
		-- & \code{title\_num}   & 68.6\perc{} &
		4.8\perc{} \\
		-- & \code{pdfid1\_num}  & 81.5\perc{} &
		2.8\perc{} \\
		-- & \code{title\_uc}    & 68.6\perc{} &
		4.8\perc{} \\
		-- & \code{pos\_eof\_min}& 100.0\perc{} &	93.4\perc{} \\
		\midrule
		+ & \code{count\_javascript}\hspace{-2mm}   & 6.0\perc{} &
		88.0\perc{} \\
		+ & \code{count\_js} & 5.2\perc{} & 83.4\perc{} \\
		+ & \code{count\_trailer} & 89.3\perc{} &
		97.7\perc{} \\
		+ & \code{pos\_page\_avg} & 100.0\perc{} &
		100.0\perc{} \\
		+ & \code{count\_endobj}  & 100.0\perc{} &
		99.6\perc{} \\
		+ & \code{createdate\_tz} & 85.5\perc{} &
		99.9\perc{} \\
		+ & \code{count\_action}  & 16.4\perc{} &
		73.8\perc{} \\
		\bottomrule
	\end{tabular}
	%\vspace{1mm}
	\label{tab:mimicus_stats}
\end{table}

\subsection{Insights on \drebin}

During the analysis of the \drebin dataset, we notice that several
benign applications are characterized by the hardware feature
\code{touchscreen}, the intent filter \code{launcher}, and the
permission \code{INTERNET}. These features frequently occur in benign
and malicious applications in the \drebin dataset and are not
particularly descriptive for benignity. Note that the interpretation
of features speaking for benign applications is challenging due to the
broader scope and the difficulty in defining benignity. We conclude
that the three features together form an artifact in the dataset that
provides an indicator for detecting benign applications.

% Still, the identified features are slightly correlated with benign
% applications and do not capture benign semantics. %As a consequence,
% an attacker could simply add these features to malicious
% applications for impeding a correct detection.

For malicious Android applications, the situation is different: The
explanation methods return highly relevant features that can be linked
to the functionality of the malware. For instance, the requested
permission \code{SEND\_SMS} or features related to accessing sensitive
information, such as the permission \code{READ\_PHONE\_STATE} and the
API call \code{getSimCountryIso}, receive consistently high scores in
our investigartion. These features are well in line with common
malware for Android, such as the \emph{FakeInstaller}
family~\citep{web:FakeInstallerL}, which is known to obtain money from
victims by secretly sending text messages~(SMS) to premium
services. Our analysis shows that the MLP network employed in \drebin
has captured indicative features directly related to the underlying
malicious activities.

\subsection{Insights on \vuldee}

% Two different explanations for the \vuldee dataset are depicted in
% \fig{fig:vuldee} in~\sect{sec:measures}.
In contrast to the datasets considered before, the features processed
by \vuldee resemble lexical tokens and are strongly interconnected on
a syntactical level. This becomes apparent in the explanations of the
method \igLong in \fig{fig:vuldee}, where adjacent tokens have mostly
equal colors. Moreover, \colorcon and \colorpro colored features in
the explanation are often separated by tokens with no color,
indicating a gradual separation of positive and negative relevance
values.

During our analysis, we notice that it is still difficult for a human
analyst to benefit from the highlighted tokens. First, an analyst
interprets the source code rather than the extracted tokens and thus
maintains a different view on the data. In \fig{fig:vuldee}, for
example, the interpretation of the highlighted \code{INT0} and
\code{INT1} tokens as buffer sizes of \num{50} and \num{100} wide
characters is misleading, since the neural network is not aware of
this relation. Second, \vuldee truncates essential parts of the
code. In \fig{fig:vuldee}, during the initialization of the
destination buffer, for instance, only the size remains as part of the
input.  Third, the large amount of highlighted tokens like semicolons,
brackets, and equality signs seems to indicate that \vuldee overfits to
the training data at hand.

Given the truncated program slices and the seemingly unrelated tokens
marked as relevant, we conclude that the \vuldee system might benefit
from extending the learning strategy to longer sequences and cleansing
the training data to remove artifacts that are irrelevant for
vulnerability discovery.

\subsection{Insights on \damd}

Finally, we consider Android applications from the \damd
dataset. Due to the difficulty of analyzing raw Dalvik bytecode, we
guide our analysis of the dataset by inspecting malicious applications
from three popular Android malware families:
GoldDream~\citep{web:GoldDream}, DroidKungFu~\citep{web:DroidKungFu1},
and DroidDream~\citep{web:DroidDream}. These families exfiltrate
sensitive data and run exploits to take full control of the device.

In our analysis of the Dalvik bytecode, we benefit from the sparsity
of the explanations from \lrp and \igShort as explained in
\sect{sec:sparsity}. Analyzing all relevant features becomes
tractable with moderate effort using these methods and we are able to
investigate the opcodes with the highest relevance in detail. We
observe that the relevant opcode sequences are linked to the
malicious functionality.% of the three malware families.

As an example, \tab{tab:casestudy-damd} depicts the opcode sequence,
that is found in all samples of the GoldDream
family.%\footnote{\eg~\hash{a4184a7fcaca52696f3d1c6cf8ce3785}}.
Taking a
closer look, this sequence occurs in the \code{onReceive} method of
the \code{com.GoldDream.zj.zjReceiver} class. In this function, the
malware intercepts incoming SMS and phone calls and stores the
information in local files before sending them to an external server.
Similarly, we can interpret the explanations of the other two malware
families, where functionality related to exploits and persistent
installation is highlighted in the Dalvik opcode sequences.

%
%  members of the DroidDream family are known to root
% infected device by running different exploits. If the attack has been
% successful, the malware installs itself as a service with the name
% \code{com.android.root.Setting}~\citep{web:DroidDream}. The top-ranked
% features determined by \igShort indeed lead to two methods of this
% very service, namely \code{com.android.root.Setting.getRawResource()}
% together with \code{com.android.root.Setting.cpFile()}. Likewise, the
% highest ranked opcode sequence of the DroidKungFu family point to the
% class, in which the decryption routine for the root exploit is stored
% in.
%
For all members of each malware family, the opcode sequences
identified using the explanation methods \lrp and \igShort are
identical, which demonstrates that the CNN in the \damd system has
learned an discriminative pattern from the underlying opcode
representation.

\section{Conclusion}
\label{sec:conclusion}

The increasing application of deep learning in security renders means
for explaining their decisions vitally important. While there exist a
wide range of explanation methods from the area of computer vision and
machine learning, it has been unclear which of these methods are
suitable for security systems. We have addressed this problem and
propose evaluation criteria that enable a practitioner to compare and
select explanation methods in the context of security. While the
importance of these criteria depends on the particular security task,
we find that the methods \igLong and \lrp comply best with all
requirements. Hence, we generally recommend these methods for
explaining predictions in security systems.

Aside from our evaluation of explanation methods, we reveal problems
in the general application of deep learning in security.  For all
considered systems under test, we identify artifacts that
substantially contribute to the overall prediction, but are unrelated
to the security task. Several of these artifacts are rooted in
peculiarities of the data. It is likely that the employed neural
networks overfit the data rather than solving the underlying task. We
thus conclude that explanations need to become an integral part of any
deep learning system to identify artifacts in the training data and to
keep the learning focused on the targeted security problem.

%\subsection{Implementations and Datasets}

Our study is a first step for integrating explainable learning in
security systems. We hope to foster a series of research that applies
and extends explanation methods, such that deep learning becomes more
transparent in computer security. To support this development, we make
all our implementations and datasets publicly available.
%This includes the four re-implemented security systems, the six
%explanation methods,and all generated explanations.
%\begin{center}
%	Project website: \webpage
%\end{center}

\section*{Acknowledgements}
\noindent The authors gratefully acknowledge funding from the German
Federal Ministry of Education and Research (BMBF) under the projects
VAMOS (FKZ 16KIS0534) and BIFOLD (FKZ 01IS18025B). Furthermore, the
authors acknowledge funding by the Deutsche Forschungsgemeinschaft
(DFG, German Research Foundation) under Germany's Excellence Strategy
EXC 2092 CASA-390781972.

\bibliographystyle{abbrvnat}
\bibliography{bib/other,%
	bib/sectubs/sec,%
	bib/sectubs/learn,%
	bib/sectubs/sectubs}

\clearpage
\appendix

\newcommand{\eqspace}{\\[-0mm]}

\subsection{Related Concepts}
\label{sec:relat-other-conc}

Some of the considered explanations methods share similarities with
techniques of adversarial examples and feature selection. While these
similarities result from an analogous analysis of the prediction
function $f_N$, the underlying objectives are fundamentally different
from explainable learning and cannot be transferred easily. In the
following, we briefly highlight these different objectives:

\subsubsection{Adversarial examples} Adversarial examples are
constructed by determining a minimal perturbation $\delta$ such that
$f_N(\xv+\delta)\neq f_N(\xv)$ for a given neural network $N$ and an
input vector \xv~\citep{CarWag17, PapMcDSinWel+18}. The perturbation
$\delta$ encodes which features need to be \emph{modified} to change
the prediction. However, the perturbation does not explain \emph{why}
$\xv$ was given the label~$y$ by the neural network. The \gradient
explanation method described in \sect{sec:methods} shares similarities
with some attacks generating adversarial examples, as the gradient
$\partial f_N/\partial x_i$ is used to quantify the difference of
$f_N$ when changing a feature $x_i$ slightly. Still, algorithms for
determining adversarial examples are insufficient for computing
reasonable explanations.

Note that we deliberately do not study adversarial examples in this
paper. Techniques for attacking and defending learning algorithms are
orthogonal to our work. These techniques can be augmented using
explanations, yet it is completely open how this can be done in a
secure manner. Recent defenses for adversarial examples based on
explanations have proven to be totally ineffective~\citep{Car19}.

\subsubsection{Feature selection} This concept aims at reducing the
dimensionality of a learning problem by selecting a subset of
discriminative features \citep{DudHarSto00}. At a first glance,
features determined through feature selection seem like a good fit for
explanation. While the selected features can be investigated and often
capture characteristics of the underlying data, they are determined
independent from a particular learning model. As a result, feature
selection methods cannot be direclty applied for explaining the
decision of a neural network.

\subsection{Incompatible Explanation Methods}
\label{sec:other-expl-meth}

Several explanation methods are not suitable for general application
in security, as they do not support common architectures of neural
networks used in this area (see Table~\ref{tab:overview-methods}). We
do not consider these methods in our evaluation, yet for completeness
we provide a short overview of these methods in the following.

\subsubsection{PatternNet and PatternAttribution}
These \wbox methods are inspired by the explanation of linear models.
While PatternNet determines gradients and replaces neural network
weights by so-called \emph{informative directions}, PatternAttribution
builds on the \lrp framework and computes explanations relative to
so-called root points whose output are~\num{0}.
Both approaches are restricted to feed-forward and convolutional
networks. Recurrent neural networks are not supported.

\subsubsection{DeConvNet and GuidedBackProp}
These methods aim at reconstructing an input $\xv$ given output $y$,
that is, mapping~$y$ back to the input space. To this end, the authors
present an approach to revert the computations of a convolutional
layer followed by a rectified linear unit~(ReLu) and max-pooling,
which is the essential sequence of layers in neural networks for image
classification. Similar to \lrp and DeepLift, both methods perform a
backwards pass through the network.
The major drawback of these methods is again the restriction to
convolutional neural networks.

\subsubsection{CAM, GradCAM, and GradCAM++}
These three \wbox methods compute relevance scores by accessing the
output of the last convolutional layer in a CNN and performing global
average pooling.  Given the activations $a_{ki}$ of the $k$-th channel
at unit $i$, GradCam learn weights $w_k$ such that
$$y \approx \sum_i\sum_k w_k a_{ki}.$$\eqspace
That is, the classification is modeled as a linear combination of the
activations of the last layer of all channels and finally
\mbox{$r_i = \sum_k w_k a_{ki}$}.
GradCam and GradCam++ extend this approach by including specific
gradients in this calculation.
All three methods are only applicable if the neural network uses a
convolutional layer as the final layer. While this setting is common
in image recognition, it is rarely used in security applications and
thus we do not analyze these methods.

\subsubsection{RTIS and MASK}
These methods compute relevance scores by solving an optimization
problem for a \emph{mask} $m$. A mask~$m$ is applied to $\xv$ as
$m\circ \xv$ in order to affect $\xv$, for example by setting features
to zero.  To this end, \citet{FonVed17} propose the optimization problem
$$
m^* = \argmin_{m\in[0,1]^d} \lambda
\lVert\mathbf{1}-m\rVert_1 +
f_N(m\circ \xv),
%\label{eq:MASK}
$$\eqspace
which determines a sparse mask, that identifies relevant features
of~$\xv$.  This can be solved using gradient descent, which thus makes
these \wbox approaches. However, solving the equation above often
leads to noisy results which is why RTIS and MASK add additional terms
to achieve smooth solutions using regularization and blurring. These
concepts, however, are only applicable for images and cannot be
transferred to other types of features.

\subsubsection{Quantitative Input Influence}
This method is another \bbox approach which calculates relevances by
changing input features and calculating the difference between the
outcomes. Let $X_{-i}U_i$ be the random variable with the $i$th input
of $X$ being replaced by a random value that is drawn from the
distribution of feature $x_i$. Then the relevance of feature $i$ for a
classification to class $c$ is given~by
$$r_i = \mathbb{E}\big[f_N(X_{-i}U_i) \neq c| X=x\big].$$\eqspace
However, when the features of $X$ are binary like in some of our
datasets this equation becomes
$$
r_i =
\begin{cases}
1\qquad f_N(x_{\neg i}) \neq c\\
0\qquad\text{else}
\end{cases}
$$\eqspace

As noted by \citet{DatSenSic16} this results in many features receiving
a relevance of zero which has no meaning. We notice that even the
extension to sets proposed by \citet{DatSenSic16} does not solve this
problem since it is highly related to degenerated explanations as
discussed in \sect{sec:availability}.

\subsection{Completeness of Datasets: Example calculation}
\label{sec:completeness_example}
In \sect{sec:availability} we discussed the problem of incomplete or
degenerated explanations from \bbox methods that can occur when there
are not enough labels from the opposite class in the perturbations. Here
we give an concrete example when enforcing \perc{5} of the labels to be
from the opposite class.

\tab{tab:perturbations2} shows the results of this experiment.
On average, \perc{29} of the samples cannot be
explained well, as the computed perturbations contain too few
instances from the opposite class. In particular, we observe that
creating malicious perturbations from benign samples is a hard problem
in the case of \drebin and \damd, where only~\perc{32.6} and
\perc{2.8} of the benign samples achieve sufficient perturbations from
the opposite class.
\begin{table}[htbp]
	\caption{Incomplete explanations of \bbox methods.
	First two columns: Samples remaining when
	enforcing at least \perc{5} perturbations of
	opposite class.}
	\centering\tablesize
	\newcommand{\mymidrule}{\cmidrule(lr){1-1}\cmidrule(lr){2-3}\cmidrule(lr){4-4}}
	\begin{tabular}{
		l
		S[table-format=2.1,table-space-text-post={\perc{}}]
		S[table-format=2.1,table-space-text-post={\perc{}}]
		S[table-format=2.1,table-space-text-post={\perc{}},table-text-alignment=right]
		}
		\toprule
		{\bfseries System \hspace{10mm}} &
		{\bfseries Class-  } &
		{\bfseries Class+ } &
		{\bfseries Incomplete}\\
		%\midrule
		\mymidrule
		\drebin    &  24.2\perc{} &  97.1\perc{} &  66.3\perc{}\\
		\mimicus   &  73.5\perc{} &  98.9\perc{} &  14.2\perc{}\\
		\vuldee    &  90.5\perc{} &  99.8\perc{} &   7.1\perc{}\\
		\damd      &   5.9\perc{} &  94.8\perc{} &  44.9\perc{}\\
		\mymidrule
		Average & 48.3\perc{} & 97.7\perc{} & \perc{33.15}\\
		\bottomrule
	\end{tabular}
	\label{tab:perturbations2}
\end{table}

\end{document}

%% file: tables/casestudy-vuldee-intro-code.tex
\begin{codelst}[style=mycppstyle]
c = split(arg[i],"=",&n);
block_flgs = strcpy((xmalloc(strlen(c[1]) + 1)),c[1]);
\end{codelst}

%% file: tables/intro-lrp.tex
\begin{codelst}
(*@\hlc[pro!18]{VAR0}@*) (*@\hlc[pro!30]{=}@*) (*@\hlc[pro!3]{FUN0}@*) (*@\hlc[pro!3]{(}@*) (*@\hlc[pro!6]{VAR1}@*) (*@\hlc[con!14]{[}@*) (*@\hlc[pro!16]{VAR2}@*) (*@\hlc[con!45]{]}@*) (*@\hlc[con!18]{,}@*) (*@\hlc[pro!100]{STR0}@*) (*@\hlc[con!44]{,}@*) (*@\hlc[con!13]{\&}@*) (*@\hlc[pro!15]{VAR3}@*) (*@\hlc[pro!1]{)}@*) (*@\hlc[con!12]{;}@*)
(*@\hlc[pro!12]{VAR0}@*) (*@\hlc[con!6]{=}@*) (*@\hlc[con!89]{strcpy}@*) (*@\hlc[pro!18]{(}@*) (*@\hlc[pro!13]{(}@*) (*@\hlc[con!16]{FUN0}@*) (*@\hlc[pro!16]{(}@*) (*@\hlc[pro!38]{strlen}@*) (*@\hlc[pro!6]{(}@*) (*@\hlc[con!0]{VAR1}@*) (*@\hlc[con!12]{[}@*) (*@\hlc[pro!34]{INT0}@*) (*@\hlc[con!21]{]}@*) (*@\hlc[pro!14]{)}@*) (*@\hlc[pro!31]{+}@*) (*@\hlc[pro!1]{INT0}@*) (*@\hlc[pro!1]{)}@*) (*@\hlc[pro!0]{)}@*) (*@\hlc[con!1]{,}@*) (*@\hlc[con!0]{VAR1}@*) (*@\hlc[pro!0]{[}@*) (*@\hlc[pro!1]{INT0}@*) (*@\hlc[con!0]{]}@*) (*@\hlc[pro!0]{)}@*) (*@\hlc[con!2]{;}@*)
\end{codelst}

%% file: tables/intro-lemna.tex
\begin{codelst}
(*@\hlc[pro!67]{VAR0}@*) (*@\hlc[pro!62]{=}@*) (*@\hlc[pro!60]{FUN0}@*) (*@\hlc[pro!58]{(}@*) (*@\hlc[pro!56]{VAR1}@*) (*@\hlc[pro!54]{[}@*) (*@\hlc[pro!53]{VAR2}@*) (*@\hlc[pro!52]{]}@*) (*@\hlc[pro!51]{,}@*) (*@\hlc[pro!51]{STR0}@*) (*@\hlc[pro!50]{,}@*) (*@\hlc[pro!49]{\&}@*) (*@\hlc[pro!49]{VAR3}@*) (*@\hlc[pro!49]{)}@*) (*@\hlc[pro!49]{;}@*)
(*@\hlc[pro!49]{VAR0}@*) (*@\hlc[pro!50]{=}@*) (*@\hlc[pro!52]{strcpy}@*) (*@\hlc[pro!96]{(}@*) (*@\hlc[pro!98]{(}@*) (*@\hlc[pro!99]{FUN0}@*) (*@\hlc[pro!100]{(}@*) (*@\hlc[pro!100]{strlen}@*) (*@\hlc[pro!100]{(}@*) (*@\hlc[pro!99]{VAR1}@*) (*@\hlc[pro!87]{[}@*) (*@\hlc[pro!86]{INT0}@*) (*@\hlc[pro!84]{]}@*) (*@\hlc[pro!83]{)}@*) (*@\hlc[pro!82]{+}@*) (*@\hlc[pro!81]{INT0}@*) (*@\hlc[pro!80]{)}@*) (*@\hlc[pro!79]{)}@*) (*@\hlc[pro!78]{,}@*) (*@\hlc[pro!77]{VAR1}@*) (*@\hlc[pro!75]{[}@*) (*@\hlc[pro!74]{INT0}@*) (*@\hlc[pro!54]{]}@*) (*@\hlc[pro!53]{)}@*) (*@\hlc[pro!52]{;}@*)
\end{codelst}

%% file: tables/intro-lime.tex
\begin{codelst}
(*@\hlc[pro!57]{VAR0}@*) (*@\hlc[con!42]{=}@*) (*@\hlc[con!6]{FUN0}@*) (*@\hlc[pro!100]{(}@*) (*@\hlc[con!12]{VAR1}@*) (*@\hlc[pro!16]{[}@*) (*@\hlc[con!34]{VAR2}@*) (*@\hlc[con!43]{]}@*) (*@\hlc[con!69]{,}@*) (*@\hlc[pro!75]{STR0}@*) (*@\hlc[pro!7]{,}@*) (*@\hlc[pro!8]{\&}@*) (*@\hlc[con!59]{VAR3}@*) (*@\hlc[pro!24]{)}@*) (*@\hlc[pro!8]{;}@*)
(*@\hlc[con!29]{VAR0}@*) (*@\hlc[con!4]{=}@*) (*@\hlc[con!78]{strcpy}@*) (*@\hlc[con!5]{(}@*) (*@\hlc[con!22]{(}@*) (*@\hlc[con!6]{FUN0}@*) (*@\hlc[pro!18]{(}@*) (*@\hlc[pro!35]{strlen}@*) (*@\hlc[pro!54]{(}@*) (*@\hlc[pro!43]{VAR1}@*) (*@\hlc[pro!37]{[}@*) (*@\hlc[pro!23]{INT0}@*) (*@\hlc[con!27]{]}@*) (*@\hlc[pro!30]{)}@*) (*@\hlc[pro!36]{+}@*) (*@\hlc[pro!53]{INT0}@*) (*@\hlc[pro!59]{)}@*) (*@\hlc[pro!88]{)}@*) (*@\hlc[con!32]{,}@*) (*@\hlc[pro!36]{VAR1}@*) (*@\hlc[pro!1]{[}@*) (*@\hlc[pro!69]{INT0}@*) (*@\hlc[con!31]{]}@*) (*@\hlc[pro!24]{)}@*) (*@\hlc[con!15]{;}@*)
\end{codelst}

%% file: tables/casestudy-vuldee-code.tex
\begin{codelst}[style=mycppstyle]
data = NULL;
data = new wchar_t[50];
data[0] = L'\\0';
wchar_t source[100];
wmemset(source, L'C', 100-1);
source[100-1] = L'\\0';
memmove(data, source, 100*sizeof(wchar_t));
\end{codelst}

%% file: tables/casestudy-vuldee-igShort.tex
\begin{codelst}
(*@\hlc[pro!34]{INT0}@*) (*@\hlc[pro!2]{]}@*) (*@\hlc[con!0]{;}@*)
(*@\hlc[pro!1]{VAR0}@*) (*@\hlc[con!32]{[}@*) (*@\hlc[con!40]{INT0}@*) (*@\hlc[con!6]{]}@*) (*@\hlc[pro!56]{=}@*) (*@\hlc[pro!100]{STR0}@*) (*@\hlc[pro!5]{;}@*)
(*@\hlc[con!18]{wchar\_t}@*) (*@\hlc[con!2]{VAR0}@*) (*@\hlc[con!9]{[}@*) (*@\hlc[con!41]{INT0}@*) (*@\hlc[pro!17]{]}@*) (*@\hlc[pro!16]{;}@*)
(*@\hlc[pro!45]{wmemset}@*) (*@\hlc[pro!1]{(}@*) (*@\hlc[con!36]{VAR0}@*) (*@\hlc[con!11]{,}@*) (*@\hlc[con!28]{STR0}@*) (*@\hlc[con!2]{,}@*) (*@\hlc[con!7]{INT0}@*) (*@\hlc[con!40]{-}@*) (*@\hlc[con!34]{INT1}@*) (*@\hlc[pro!9]{)}@*) (*@\hlc[pro!4]{;}@*)
(*@\hlc[con!9]{VAR0}@*) (*@\hlc[pro!11]{[}@*) (*@\hlc[pro!13]{INT0}@*) (*@\hlc[pro!10]{-}@*) (*@\hlc[pro!17]{INT1}@*) (*@\hlc[pro!4]{]}@*) (*@\hlc[pro!8]{=}@*) (*@\hlc[pro!48]{STR0}@*) (*@\hlc[con!0]{;}@*)
(*@\hlc[con!6]{memmove}@*) (*@\hlc[pro!4]{(}@*) (*@\hlc[pro!1]{VAR0}@*) (*@\hlc[pro!10]{,}@*) (*@\hlc[pro!3]{VAR1}@*) (*@\hlc[pro!10]{,}@*) (*@\hlc[con!13]{INT0}@*) (*@\hlc[con!7]{*}@*) (*@\hlc[pro!3]{sizeof}@*) (*@\hlc[pro!6]{(}@*) (*@\hlc[con!4]{wchar\_t}@*) (*@\hlc[pro!7]{)}@*) (*@\hlc[pro!9]{)}@*) (*@\hlc[pro!4]{;}@*)
\end{codelst}

%% file: tables/casestudy-vuldee-lime.tex
\begin{codelst}
(*@\hlc[pro!13]{INT0}@*) (*@\hlc[con!1]{]}@*) (*@\hlc[pro!19]{;}@*)
(*@\hlc[con!22]{VAR0}@*) (*@\hlc[pro!8]{[}@*) (*@\hlc[con!12]{INT0}@*) (*@\hlc[pro!11]{]}@*) (*@\hlc[pro!15]{=}@*) (*@\hlc[con!12]{STR0}@*) (*@\hlc[pro!23]{;}@*)
(*@\hlc[pro!19]{wchar\_t}@*) (*@\hlc[pro!33]{VAR0}@*) (*@\hlc[con!14]{[}@*) (*@\hlc[con!4]{INT0}@*) (*@\hlc[con!6]{]}@*) (*@\hlc[pro!2]{;}@*)
(*@\hlc[con!50]{wmemset}@*) (*@\hlc[con!6]{(}@*) (*@\hlc[pro!53]{VAR0}@*) (*@\hlc[pro!10]{,}@*) (*@\hlc[pro!5]{STR0}@*) (*@\hlc[pro!5]{,}@*) (*@\hlc[con!44]{INT0}@*) (*@\hlc[pro!28]{-}@*) (*@\hlc[pro!100]{INT1}@*) (*@\hlc[con!35]{)}@*) (*@\hlc[pro!18]{;}@*)
(*@\hlc[pro!18]{VAR0}@*) (*@\hlc[pro!29]{[}@*) (*@\hlc[con!28]{INT0}@*) (*@\hlc[pro!39]{-}@*) (*@\hlc[pro!87]{INT1}@*) (*@\hlc[pro!7]{]}@*) (*@\hlc[con!5]{=}@*) (*@\hlc[con!2]{STR0}@*) (*@\hlc[pro!6]{;}@*)
(*@\hlc[pro!45]{memmove}@*) (*@\hlc[con!16]{(}@*) (*@\hlc[con!9]{VAR0}@*) (*@\hlc[con!9]{,}@*) (*@\hlc[pro!9]{VAR1}@*) (*@\hlc[con!5]{,}@*) (*@\hlc[pro!6]{INT0}@*) (*@\hlc[con!3]{*}@*) (*@\hlc[pro!13]{sizeof}@*) (*@\hlc[con!25]{(}@*) (*@\hlc[pro!19]{wchar\_t}@*) (*@\hlc[con!31]{)}@*) (*@\hlc[con!21]{)}@*) (*@\hlc[pro!8]{;}@*)
\end{codelst}

%% file: tables/casestudy-damd-shortened.tex
\begin{tabular}{
  cl@{\hskip 0.2in}l
}
  \toprule
  \bfseries Id & \bfseries \lrp & \bfseries \lemna\\
  \midrule
  0 & \cellcolor{pro!7} \ttfamily invoke-virtual & \cellcolor{con!11} \ttfamily invoke-virtual \\\\[ -3.3mm]
  1 & \cellcolor{pro!27} \ttfamily move-result-object & \cellcolor{con!11} \ttfamily move-result-object \\\\[ -3.3mm]
  2 & \cellcolor{pro!48} \ttfamily if-eqz & \cellcolor{con!11} \ttfamily if-eqz \\\\[ -3.3mm]
  3 & \cellcolor{pro!33} \ttfamily const-string & \cellcolor{con!11} \ttfamily const-string \\\\[ -3.3mm]
  4 & \cellcolor{pro!28} \ttfamily invoke-virtual & \cellcolor{con!11} \ttfamily invoke-virtual \\\\[ -3.3mm]
  5 & \cellcolor{pro!17} \ttfamily move-result-object & \cellcolor{con!11} \ttfamily move-result-object \\\\[ -3.3mm]
  6 & \cellcolor{pro!56} \ttfamily check-cast & \cellcolor{con!11} \ttfamily check-cast \\\\[ -3.3mm]
  7 & \cellcolor{pro!99} \ttfamily array-length & \cellcolor{con!11} \ttfamily array-length \\\\[ -3.3mm]
  8 & \cellcolor{pro!48} \ttfamily new-array & \cellcolor{con!11} \ttfamily new-array \\\\[ -3.3mm]
  9 & \cellcolor{pro!13} \ttfamily const/4 & \cellcolor{con!11} \ttfamily const/4 \\\\[ -3.3mm]
  10 & \cellcolor{pro!5} \ttfamily array-length & \cellcolor{con!11} \ttfamily array-length \\\\[ -3.3mm]
  11 & \cellcolor{pro!14} \ttfamily if-ge & \cellcolor{con!11} \ttfamily if-ge \\\\[ -3.3mm]
  12 & \cellcolor{pro!6} \ttfamily aget-object & \cellcolor{con!10} \ttfamily aget-object \\\\[ -3.3mm]
  \bottomrule
\end{tabular}

%% file: tables/casestudy-drebin-incomplete.tex
\begin{tabular}{
  cll
}
  \toprule
  \bfseries Id & \bfseries \gradient & \bfseries \shap\\
  \midrule
  0 & \cellcolor{pro!100} \ttfamily feature::android.hardware.touchscreen & \cellcolor{pro!0} \ttfamily feature::android.hardware.touchscreen \\\\[ -3.2mm]
  1 & \cellcolor{pro!61} \ttfamily intent::android.intent.category.LAUNCHER & \cellcolor{pro!0} \ttfamily intent::android.intent.category.LAUNCHER \\\\[ -3.2mm]
  2 & \cellcolor{pro!26} \ttfamily real\_permission::android.permission.INTERNET & \cellcolor{pro!0} \ttfamily real\_permission::android.permission.INTERNET \\\\[ -3.2mm]
  3 & \cellcolor{pro!18} \ttfamily api\_call::android/webkit/WebView & \cellcolor{pro!0} \ttfamily api\_call::android/webkit/WebView \\\\[ -3.2mm]
  4 & \cellcolor{pro!2} \ttfamily intent::android.intent.action.MAIN & \cellcolor{pro!0} \ttfamily intent::android.intent.action.MAIN \\\\[ -3.2mm]
  5 & \cellcolor{pro!1} \ttfamily url::translator.worldclockr.com & \cellcolor{pro!0} \ttfamily url::translator.worldclockr.com \\\\[ -3.2mm]
  6 & \cellcolor{con!0} \ttfamily url::translator.worldclockr.com/android.html & \cellcolor{pro!0} \ttfamily url::translator.worldclockr.com/android.html \\\\[ -3.2mm]
  7 & \cellcolor{con!3} \ttfamily permission::android.permission.INTERNET & \cellcolor{pro!0} \ttfamily permission::android.permission.INTERNET \\\\[ -3.2mm]
  8 & \cellcolor{con!19} \ttfamily activity::.Main & \cellcolor{pro!0} \ttfamily activity::.Main \\\\[ -3.2mm]
  \bottomrule
\end{tabular}

%% file: tables/casestudy-mimicus-no-stability-shortened.tex
\begin{tabular}{
  cl@{\hskip 0.2in}l
}
  \toprule
  \bfseries Id & \bfseries \lemna (Run 1) & \bfseries \lemna (Run 2)\\
  \midrule
  0 & \cellcolor{pro!100} \ttfamily pos\_page\_min & \cellcolor{pro!4} \ttfamily pos\_page\_min \\\\[ -3.3mm]
  1 & \cellcolor{pro!94} \ttfamily count\_js & \cellcolor{pro!56} \ttfamily count\_js \\\\[ -3.3mm]
  2 & \cellcolor{pro!77} \ttfamily count\_javascript & \cellcolor{pro!22} \ttfamily count\_javascript \\\\[ -3.3mm]
  3 & \cellcolor{pro!72} \ttfamily pos\_acroform\_min & \cellcolor{pro!35} \ttfamily pos\_acroform\_min \\\\[ -3.3mm]
  4 & \cellcolor{pro!69} \ttfamily ratio\_size\_page & \cellcolor{pro!66} \ttfamily ratio\_size\_page \\\\[ -3.3mm]
  5 & \cellcolor{pro!66} \ttfamily pos\_image\_min & \cellcolor{pro!12} \ttfamily pos\_image\_min \\\\[ -3.3mm]
  6 & \cellcolor{pro!66} \ttfamily count\_obj & \cellcolor{con!5} \ttfamily count\_obj \\\\[ -3.3mm]
     & \cellcolor{pro!0} \ttfamily ... & \cellcolor{con!0} \ttfamily ... \\\\[ -3.3mm]
  27 & \cellcolor{pro!16} \ttfamily pos\_image\_max & \cellcolor{pro!17} \ttfamily pos\_image\_max \\\\[ -3.3mm]
  28 & \cellcolor{pro!15} \ttfamily count\_page & \cellcolor{pro!78} \ttfamily count\_page \\\\[ -3.3mm]
  29 & \cellcolor{pro!14} \ttfamily len\_stream\_avg & \cellcolor{pro!72} \ttfamily len\_stream\_avg \\\\[ -3.3mm]
  30 & \cellcolor{pro!12} \ttfamily pos\_page\_avg & \cellcolor{pro!51} \ttfamily pos\_page\_avg \\\\[ -3.3mm]
  31 & \cellcolor{pro!10} \ttfamily count\_stream & \cellcolor{pro!18} \ttfamily count\_stream \\\\[ -3.3mm]
  32 & \cellcolor{con!9} \ttfamily moddate\_tz & \cellcolor{con!4} \ttfamily moddate\_tz \\\\[ -3.3mm]
  33 & \cellcolor{con!65} \ttfamily len\_stream\_max & \cellcolor{con!23} \ttfamily len\_stream\_max \\\\[ -3.3mm]
  34 & \cellcolor{con!100} \ttfamily count\_endstream & \cellcolor{pro!7} \ttfamily count\_endstream \\\\[ -3.3mm]
  \bottomrule
\end{tabular}